\documentclass{article}

\usepackage{microtype}
\usepackage{graphicx}
\usepackage{subfigure}
\usepackage{booktabs} 
\usepackage{amsmath}
\usepackage{amssymb}
\usepackage{amsmath}
\usepackage{amsthm}

\newcommand{\intMean}[1]{\langle #1 \rangle}

\newtheorem{theorem}{Theorem}[section]
\newtheorem{lemma}[theorem]{Lemma}
\newtheorem{corollary}[theorem]{Corollary}

\usepackage{xr-hyper}

\usepackage{hyperref}
    
\usepackage{tikz}
\usetikzlibrary{decorations.pathreplacing,calc}
\newcommand{\tikzmark}[1]{\tikz[overlay,remember picture] \node (#1) {};}

\newcommand*{\AddNote}[4]{%
    \begin{tikzpicture}[overlay, remember picture]
        \draw [decoration={brace,amplitude=0.5em},decorate,ultra thick]
            ($(#3)!(#1.north)!($(#3)-(0,1)$)$) --  
            ($(#3)!(#2.south)!($(#3)-(0,1)$)$)
                node [align=center, text width=0.8cm, pos=0.5, anchor=west] {#4};
    \end{tikzpicture}
}

\usepackage[accepted]{icml2021}

\def\R{\mathbb{R}}

\def\d{\mathrm{d}}
\DeclareMathOperator{\Div}{div}
\def\calL{\mathcal{L}}
\def\calA{\mathcal{A}}
\newcommand{\ps}[1]{\langle #1 \rangle}
\newcommand{\psbig}[1]{\big\langle #1 \big\rangle}
\newcommand{\norm}[1]{\| #1 \|}

\def\manifold{\Sigma}
\def\frakg{\mathfrak{g}}
\def\calM{\mathcal{M}}
\def\loss{L_X}

\def\Ld{L^2}

\def\Ldznu{L^2_0(\nu_\manifold)}
\def\fcnC{\mathcal{C}}
\def\E{\mathbb{E}}
\def\d{{\rm d}}

\icmltitlerunning{Better Training using Weight-Constrained Stochastic Dynamics} 

\begin{document}

\twocolumn[
\icmltitle{Better Training using Weight-Constrained Stochastic Dynamics}


\icmlsetsymbol{equal}{*}

\begin{icmlauthorlist}
\icmlauthor{Benedict Leimkuhler}{edM}
\icmlauthor{Tiffany Vlaar}{edM}
\icmlauthor{Timothée Pouchon}{edM}
\icmlauthor{Amos Storkey}{edI}
\end{icmlauthorlist}

\icmlaffiliation{edM}{Department of Mathematics, University of Edinburgh, United Kingdom}
\icmlaffiliation{edI}{Department of Informatics, University of Edinburgh, United Kingdom}

\icmlcorrespondingauthor{Tiffany Vlaar}{Tiffany.Vlaar@ed.ac.uk}

\icmlkeywords{Constrained Optimization, Langevin Dynamics, 
  Orthogonality Preservation.}

\vskip 0.3in
]

\printAffiliationsAndNotice{} 

\begin{abstract}
We employ constraints to control the parameter space of deep neural networks throughout training. The use of customized, appropriately designed constraints can reduce the vanishing/exploding gradients problem, improve smoothness of classification boundaries, control weight magnitudes and stabilize deep neural networks, and thus enhance the robustness of training algorithms and the generalization capabilities of neural networks. We provide a general approach to efficiently incorporate constraints into a stochastic gradient Langevin framework, allowing enhanced exploration of the loss landscape. We also present specific examples of constrained training methods motivated by orthogonality preservation for weight matrices and explicit weight normalizations. Discretization schemes are provided both for the overdamped formulation of Langevin dynamics and the underdamped form, in which momenta further improve sampling efficiency. These optimization schemes can be used directly, without needing to adapt neural network architecture design choices or to modify the objective with regularization terms, and see performance improvements in classification tasks.
\end{abstract}

\section{Introduction}
We study stochastic training methods based on Langevin dynamics combined with algebraic constraints. Our general framework allows for incorporating constraints into standard training schemes and sampling methods for neural networks. Constraints provide direct control of the parameter space of a model and hence afford a means to improve its generalization performance. As applications, we consider magnitude control and orthogonality of neural network weights. 

Current approaches to enhance the generalization performance of overparameterized neural networks consist of both explicit and implicit regularization techniques \cite{neyshabur2015}. Examples of the former are L1 \cite{LASSO2, LASSO1} and L2 \cite{ridge} regularization, which modify the loss by adding a parameter norm penalty term. Batch normalization (BatchNorm) \cite{BatchNorm} is a technique that causes an implicit regularization effect. BatchNorm can be viewed as tantamount to a constraint imposed on the network's parameters during training. Although BatchNorm is widely used, explanations for the method's success remain elusive 
\cite{BNhelpopt,PyHessian}.  The reliance on increasingly complex strategies does little to enhance the explainability of neural networks, so robust simplification of all aspects of training is desirable. The constrained approach proposed in this paper provides a conceptually straightforward and interpretable framework that offers direct control of parameter spaces, without requiring modifications to the neural network architecture or objective. The transparency of this approach allows for drawing a direct connection between the use of weight constraints and the generalisation performance of the resulting neural network. 

In neural network (NN) training one aims to minimize the loss $L_X(\theta)$ for parameters $\theta\in \mathbb{R}^{n}$ and data $X$.  Constraints can be seen as limiting cases of penalty-based regularization which replaces minimization of the loss $L_X(\theta)$ by that of the augmented loss 
$L^c_X(\theta) = L_X(\theta) + \frac{1}{\varepsilon^2}g(\theta)^2$,
where $g(\cdot)$ is a suitable smooth function of the parameters. In the limit $\varepsilon\rightarrow 0$, these penalty terms introduce an undesirable stiffness and consequent stability restriction in gradient-based training, which limits the choice of step size (see  Figure \ref{Stiffness} for an illustration). It is therefore natural to relate the above system to a constrained optimization task subject to $g(\theta)=0$ (see Section \ref{sec:constraints}).  

\begin{figure*}[!hbt]
\vskip 0.2in
\vspace*{-0.2cm}
\begin{center}
\centerline{\includegraphics[width=\textwidth]{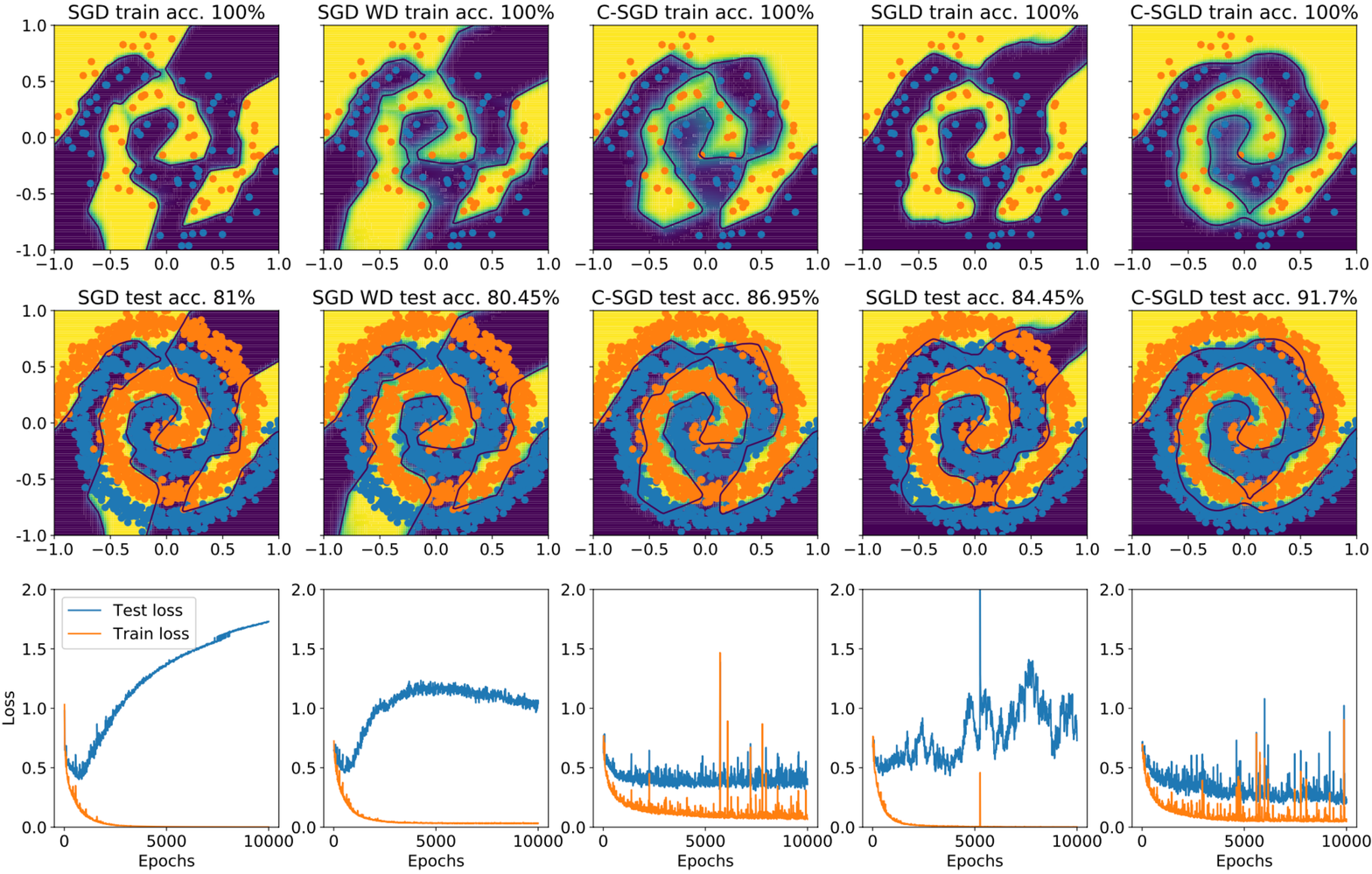}}
\vspace*{-1cm}
\caption{Classifiers obtained using different optimizers: SGD (1st column), SGD with weight decay (WD) (2nd col.), constrained SGD (C-SGD) (3rd col.), SGLD \cite{WeT11} (4th col.), constrained SGLD (5th col.) using a 500-node single hidden layer perceptron for a spiral binary classification problem. Top and middle row show training and test data points, respectively, and decision boundaries of the trained classifier. Bottom row shows loss curves. Hyperpar. settings: all: $h = 0.05$, 2\% subsampling; SGD with WD = 1e-4; C-SGD: $r_0 = 1, r_1 = 5$ 
(see Eq. \eqref{eq:circleConstraint}); SGLD and C-SGLD: $\tau$ = 5e-5 (see Eq. \eqref{eq:constrainedOverdampedLangevin}). We observe that although the use of WD can stabilize the test loss, it does not improve test accuracy (2nd col.). In contrast, our constrained approach (3rd col.) maintains a stable test loss throughout training and improved generalization performance. The use of additive noise (or low temperature) in combination with the constraints (C-SGLD, 5th col.) strongly outperforms standard SGD: 91.7\% vs. 81\% test acc., resp., and obtains smoothened classification boundaries.}
\label{overviewfig_CNet}
\end{center}
\vskip -0.2in
\end{figure*}
\begin{figure*}[ht]
\begin{center}
\centerline{\includegraphics[clip, trim=0cm 3.5cm 0cm 3.5cm,width=0.65\linewidth]{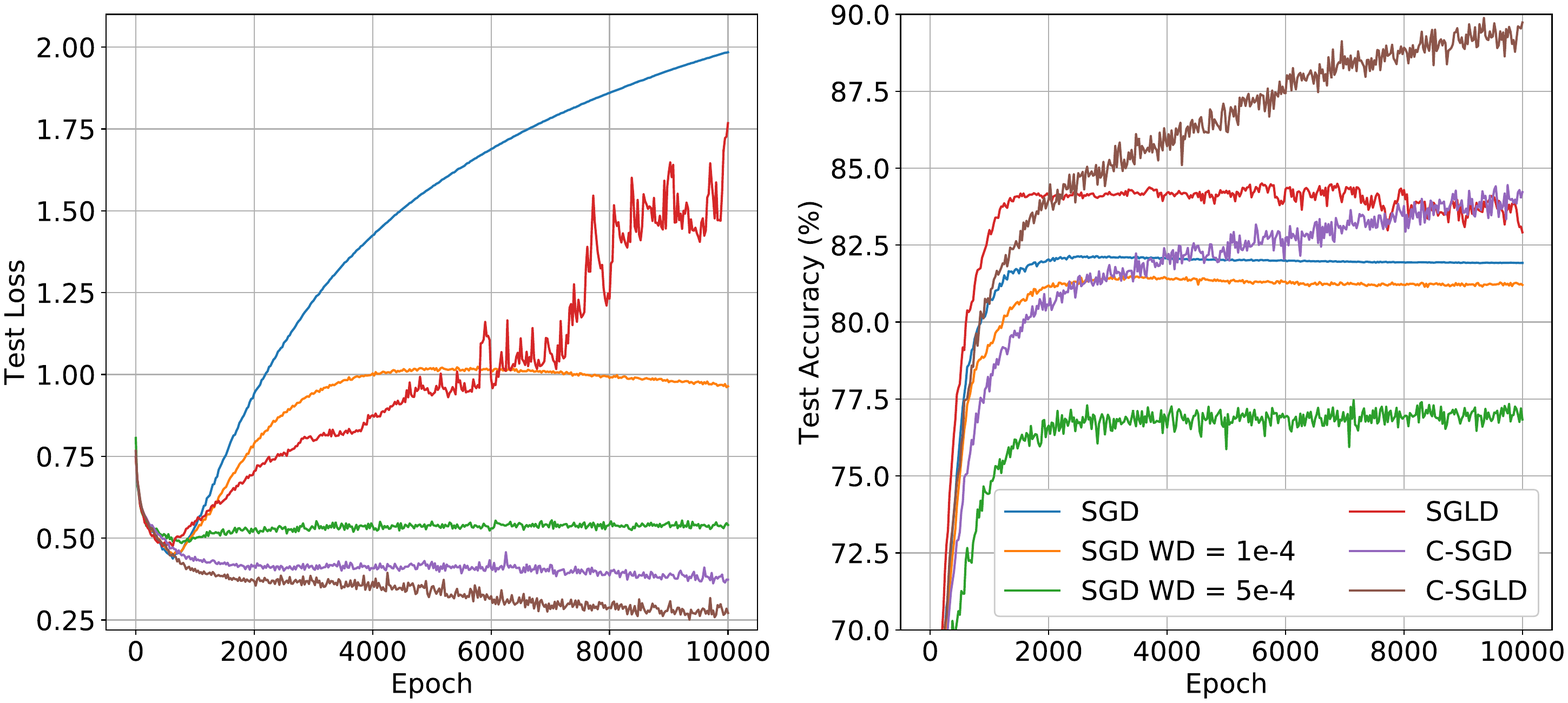}}
\vskip -0.1in
\caption{Same data and hyperparameter settings as for Fig. \ref{overviewfig_CNet}, but these results are averaged over 100 runs.
Constrained approaches, C-SGD and C-SGLD (with additive noise), clearly outperform standard SGD and SGD with WD in terms of test loss and test accuracy.  }
\label{avg_overviewfig_CNet}
\end{center}
\vskip -0.25in
\end{figure*}
\begin{figure*}[ht]
\begin{center}
\centerline{\includegraphics[clip, trim=0cm 5.7cm 0cm 5.7cm,width=0.85\linewidth]{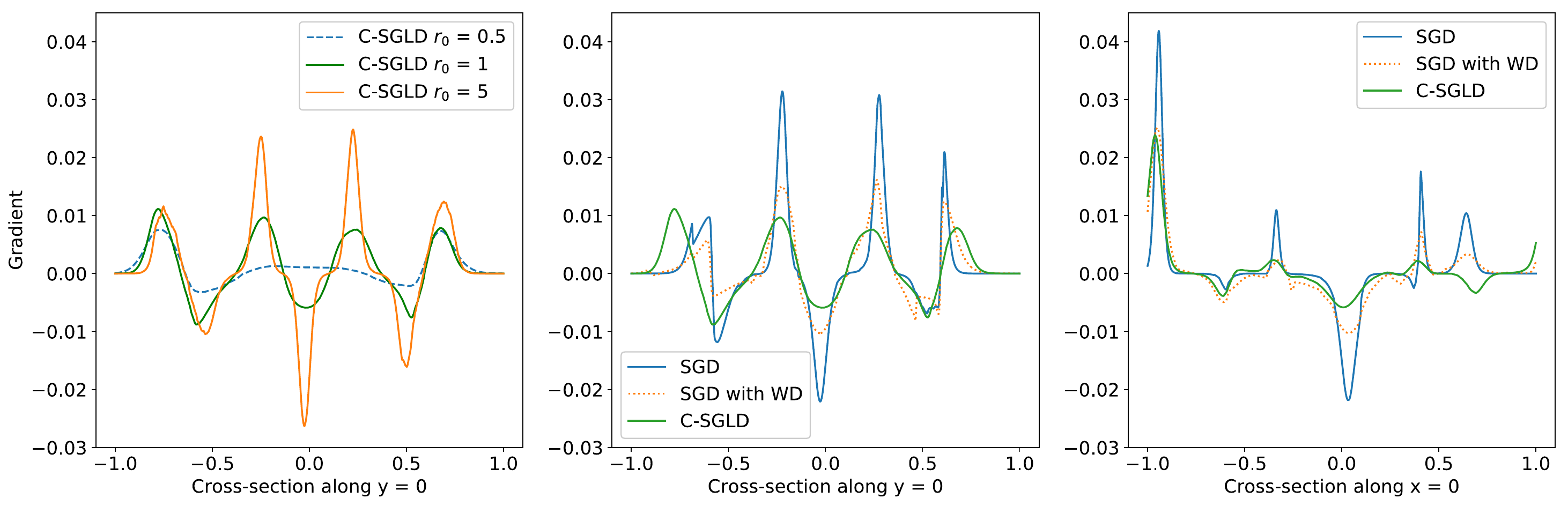}}
\vskip -0.1in
\caption{Same data and hyperparameter settings as in Fig. \ref{overviewfig_CNet} and \ref{avg_overviewfig_CNet}. We show gradients of the prediction along horizontal ($y=0$) and vertical cross-sections ($x = 0$). The results are averaged over 100 runs and evaluated at 10,000 epochs. Our constrained approach C-SGLD exhibits less sharp transitions between classes than SGD (middle/right plot). The size of the constraint directly controls this property (this is illustrated in the left plot for the input layer with constraint size $r_0$ in Eq. \eqref{eq:circleConstraint} for all input layer weights).}
\label{gradients}
\end{center}
\vskip -0.25in
\end{figure*}

A popular NN training scheme is stochastic gradient descent (SGD). SGD may be improved by incorporating momenta \cite{sutskever2013} and additive noise \cite{WeT11,coldposterior}, or more generally by embedding the loss gradient in a Langevin dynamics (LD) framework \cite{Cheng2017}. We will combine the resulting discretized stochastic differential equation (SDE) approach with constraints (Sec. \ref{sec:constrainedSDEs}). 
The benefit of using constrained SDEs for NN training is illustrated in Figure \ref{overviewfig_CNet}, where the combination of using additive noise and magnitude constraints 
(as defined by Eq. \eqref{eq:circleConstraint}) leads to smoother classification boundaries and significantly enhanced generalization performance (compare the 5th column, the constrained SDE approach, with column 1, standard SGD). These observations are maintained over 100 runs (see Fig. \ref{avg_overviewfig_CNet}, Fig. \ref{gradients}, and Table \ref{curvature}). We distinguish between two different types of smoothness of the resulting classifiers: first, the curvature of the classification boundary and second, the sharpness of the transition between prediction regions belonging to different classes. As shown in Table \ref{curvature} and Fig. \ref{gradients} the use of magnitude constraints throughout training generates classifiers which exhibit both types of smoothness. The use of additive noise throughout training further reduces the curvature of the classification boundary. In contrast, the use of weight decay is not sufficient for SGD to obtain the same levels of smoothness. See Appendix \ref{sec:NumericsAppx} for further numerical details.

 \begin{table} 
 \vskip -0.3in
\caption{Accompanies Fig. \ref{overviewfig_CNet} and \ref{avg_overviewfig_CNet}, with same hyperparameter settings. We present estimates of the mean, standard deviation (std), and maximum (max) curvature of classifier boundaries obtained using different optimizers evaluated over 100 runs after training for a fixed number of 10,000 epochs. 
We computed our curvature estimates using the method described in Appx. \ref{sec:NumericsAppx}, which we suggest is indicative of the curvature of the locally smoothed classification boundary and allows us to compare the relative curvature estimates of classifiers trained using different optimizers. The combined use of constraints and additive noise (C-SGLD) obtains much lower curvatures compared to SGD with weight decay (WD).} \label{curvature}
\center
\begin{tabular}{l|l|l|l} 
     &  \multicolumn{3}{c}{\textit{Curvature Approximation}} \\ 
      \textit{Optimizer} & Mean & Std & Max
      \\ \hline 
      SGD & 519 & 4.33 $\cdot 10^4$ & 3.26$\cdot 10^8$ \\ \hline
      SGD with WD & 51.1 & 3.80 $\cdot 10^3$ & 1.14 $\cdot 10^7$ \\ \hline
      C-SGD  & 9.38 & 317 & 5.58 $\cdot 10^{5}$ \\ \hline
      SGLD & 8.73 & 189 & 6.27 $\cdot 10^{5}$ \\ \hline
      \textbf{C-SGLD} & 6.08 & 40.8 & 1.43 $\cdot 10^5$
\end{tabular} 
\end{table}

Apart from such magnitude constraints, the general framework provided in this paper allows for straightforward incorporation of other constraints. Another specific example we consider is orthogonality of the weight matrix. We provide detailed algorithms for both of these purposes and for a general constraint in a Langevin dynamics setting (Sec. \ref{sec:constrainedSDEs} and Appendix \ref{sec:discretization}) and show improved generalization performance on classification tasks (Sec. \ref{sec:numerics}).

Concretely, our contributions are:
\begin{itemize}
    \item We introduce the use of constrained stochastic differential equations for neural network training. 
    \item We provide a general mathematical framework that allows for implementation of new constraints.
    \item We propose specific constraints, magnitude control and orthogonality of neural network weights, and provide algorithms to accompany these. The benefit of using these is illustrated for several networks and datasets and is shown to outperform soft constraints (such as weight decay or orthogonal regularization).
    \item We provide PyTorch code to support our algorithms, which can be found on \url{https://github.com/TiffanyVlaar/ConstrainedNNtraining}
\end{itemize}
 
\section{Background and Related Work}
Neural network loss landscapes are notoriously difficult to characterize rigorously due to their high-dimensionality and non-convexity. Although they appear to contain multiple, roughly equivalent local minima  which exhibit nearly zero training loss \cite{Choromanska2015,kawaguchi2016,largebatch}, 
not all these minima obtain the same generalization performance \cite{Chaudhari2017, Wu2017}. The training of deep neural networks is hypersensitive to e.g., the choice of initialization \cite{sutskever2013}, optimizer \cite{Wilson2017}, and 
hyperparameter settings \cite{Jastrzebski2018}, including learning rate scheduling \cite{SGDR,cyclicallr}. 
Without careful hyperparameter tuning, the loss landscape may not be explored sufficiently by the optimization scheme, thus resulting in a reduced generalization performance of the trained network \cite{elasticaveragingSGD,largebatch}. Sampling methods, which use small amounts of additive noise \cite{LMV,coldposterior}, have been found to enhance exploration and speed the approach to `good' minima, which enhance their generalization to nearby data sets. Hence, we incorporate the flexibility to use additive noise to enhance exploration in our optimization schemes by taking a constrained SDE approach to neural network training. We propose a general mathematical framework for this purpose and consider the ergodic properties of the idealized SDEs associated with gradient schemes, which may help these methods to ensure robust exploration of a useful range of parameters (Sec. \ref{sec:constrainedSDEs}). We further propose specific constraints (Sec. \ref{sec:constraints}) and show that the use of these leads to enhanced performance compared to soft constraints, such as weight decay or orthogonal regularization (Sec. \ref{sec:numerics}).

\textbf{Magnitude control of neural network weights.} 
In this work we consider a circle constraint, which limits the magnitude of the size of the weights (we typically leave the biases unconstrained). A corresponding soft constraint, which adds a penalty term to the loss, is weight decay or L2 regularization \cite{ridge}. We also propose a sphere constraint, which is analogous to max-norm \cite{maxnorm,dropout2} as used in some regularization procedures. However, applying this constraint in combination with additive noise does yield a distinctive training method. 

\par{\bf Orthogonality of the weight matrix.}  The concept of orthogonality has surfaced several times in the recent neural network literature. Orthogonal matrices have properties (norm preservation, unit singular values) which are thought to provide enhanced numerical stability \cite{Zhou2006, Rodriguez2017}. An orthogonal matrix $Q\in\mathbb{R}^{r\times s}$ (i.e., $Q^TQ = I_s$) is an isometry:
$\|Qz\| = \|z\|$ $\forall z\in\mathbb{R}^s$.
Orthogonal weight matrices were shown to mitigate the vanishing/exploding gradient problem in RNNs \cite{PascanuRNNs,UnitaryEvolutionRNNs,RNNsLongTermDependencies} and are developing a growing following in the CNN literature as well
\cite{Rodriguez2017, OrthReg_CNNs,CNN_stiefelmanifold,LipschitzCNNs}. Orthogonal initialization is linked to achieving dynamical isometry \cite{Saxe2013, Pennington2017, Pennington2018}, which can accelerate training. \citet{10000CNN} were able to train 10,000 layer vanilla CNNs, 
without learning rate decay, BatchNorm or residual connections, by using initial orthogonal convolution kernels. 

Methods for enforcing orthogonality during training include the use of `soft' constraints which add a restraint term to the loss \cite{NeuralPhotoEditing,BeyondGoodInit,OrthReg_CNNs} and hard constraints based on optimization over Stiefel manifolds \cite{CNN_stiefelmanifold, OrthDNN}. The latter requires repeated singular value decomposition of high-dimensional matrices during training, which is costly. 
Int his work we propose a straightforward algorithm to incorporate orthonormality constraints for rectangular matrices within our NN training framework, with manageable additional cost. We make no empirical claims over other manifold optimization methods, but rather provide a framework for network optimization that is theoretically sound, flexible enough to incorporate new constraints, and demonstrates good properties relative to standard SGD training or simple soft constraint approaches.

\textbf{Constrained SDEs.} In this work we focus on optimization schemes for neural networks using constrained Langevin dynamics in both its overdamped and underdamped (with momentum) form. A discussion of the properties of unconstrained Langevin dynamics in its overdamped and underdamped forms was studied in \citet{Pav14}. We consider the specific issues associated to the extension of the standard framework to constrained SDEs. The ergodic properties of constrained Langevin (in the absence of gradient noise) were previously studied in \citet{LSR10} (overdamped) and \citet{LRS12} (underdamped). Exponential convergence to equilibrium for constrained overdamped Langevin is a consequence of a Poincaré inequality. Poincaré inequalities on manifolds and their use in the analysis of diffusion processes are presented in \citet{BGL13}, Chapter 4. Finally, Langevin dynamics discretizations are studied in \citet{FaL09,LSR10} (overdamped) and \citet{LRS12, LeM16} (underdamped).

An alternative to the use of constrained SDEs are constrained Hamiltonian Monte Carlo (HMC) methods \cite{GS2017,ZHG2018,LSZ2020}.
Although HMC schemes have nil sampling bias if fully converged, their acceptance rates depend on stepsize and system size \cite{Beskos2013,BRSS2017}. In practice SDE-based methods are often preferred in many high-dimensional sampling calculations compared to HMC schemes as they are found to offer greater overall efficiency for a fixed computational budget.

\section{Neural Networks with Constraints} 	\label{sec:constraints}
Imposing good priors on neural networks is known to improve performance, e.g. convolutional neural networks (CNNs) suit image datasets better than overparameterized fully connected NNs, despite being a subset of the latter \cite{haystack}. Using constraints also arises naturally in the control of vanishing/exploding gradients. In Appendix \ref{sec:nnGradient} we illustrate this and also provide a connection between the magnitude of the weights and the smoothness of the interpolant. These observations suggest the use of constraints to control the magnitudes of individual weights and/or to limit the growth of gradients in deep networks. 
We present various approaches in this section.

We consider a $L$-layer neural network, which has parameters $\theta\in\mathbb{R}^{n}$, with a weight matrix $W^{\ell} \in\mathbb{R}^{d^\ell\times d^{\ell-1}}$ and bias vector $b^\ell \in\mathbb{R}^{d^{\ell}}$ for each layer $\ell$.  
To allow for inequality constraints, we define slack variables vector $\xi\in\mathbb{R}^{n^\xi}$ 
and consider variable $q = (\theta,\xi)\in\mathbb{R}^{d}$, where $d=n+n^\xi$.
The constraint manifold is 
\begin{equation}	\label{eq:constraintManifold}
\Sigma = \{ q\in\mathbb{R}^d \mid g(q)=0  \}, \ \ g:\mathbb{R}^d\to\mathbb{R}^m.
\end{equation}
We partition $\theta = (\theta^u,\theta^c)$ into unconstrained $\theta^u\in\mathbb{R}^{n^u}$ and constrained $\theta^c\in\mathbb{R}^{n^c}$ parameters. We typically only constrain the neural network weights, not the biases.
\par{\bf Circle constraints:}
In a {\em circle constraint}, we restrict each parameter in $\theta^c$ as 
$|\theta^c_i|\leq r_i$, where $r_i>0$ is given.
We thus introduce $m = n^c=n^\xi$ slack variables $\xi_i$
and define
\begin{equation}	\label{eq:circleConstraint}
g_i(q) = |\theta^c_i|^2 + |\xi_i|^2 - r_i^2 \qquad 1\leq i\leq m.
\end{equation}
If $q\in\Sigma$, then the parameters in $\theta^c$ are bounded as desired.

\par{\bf Sphere constraints:} In a similar way, we could opt to restrict the sums of squares of weights associated to the input channels of any node. 
For layer $\ell$, we denote the $i$-th row of the weight matrix $W^{\ell}$ as $\theta^{c,i}$, set $\theta^u=b^\ell$, 
 introduce $m = d^{\ell}$ slack variables $\xi_i$, 
and define as \textit{sphere constraint}: 
\begin{align}
    g_i(q) = \|\theta^{c,i}\|^2+|\xi_i|^2-r^2_i, \ 1 \leq i \leq m, \label{sphereconstraint}
\end{align}
where $\| \cdot \|$ denotes the Euclidean norm. Sphere constraints are analogous to max-norm \cite{maxnorm,dropout2}, but have been unexplored in combination with additive noise. We leave this for future work. 

\par{\bf Orthogonality constraints:}  
We set $\theta^u = b^\ell$, 
and define as \textit{orthogonality constraint} for layer $\ell$ with $n^{\ell}$ parameters 
\begin{equation}	\label{eq:orthogonalConstraint}
g(q) =
\left\{
\begin{array}{ll}
\big(W^\ell\big)^T W^\ell - I_{n^{\ell-1}} 	&\text{if }n^{\ell-1}\leq n^{\ell}, \\
W^\ell \big(W^\ell\big)^T  - I_{n^{\ell}} 	&\text{otherwise}.
\end{array}
\right.
\end{equation}
As the matrix equality $g(q) = 0$ is symmetric, it corresponds to $m = s(s + 1)/2$ constraints, where
$s$ = min$\{ n^{l-1}, n^l \}$.

\section{Constrained SDEs and their Discretization}	\label{sec:constrainedSDEs}	
In this chapter we describe SDE-based methods for constrained neural network training. We first introduce standard (unconstrained) Langevin dynamics in Section \ref{subsec:Langevin}. Then in Section \ref{subsec:constrainedLangevin} we discuss properties of constrained Langevin dynamics (LD), such as ergodicity and exponential convergence to equilibrium, which ensures the effectiveness of our schemes as training methods. In Section \ref{subsec:discretization} we discuss the discretization of constrained Langevin dynamics in both the overdamped and the underdamped case, where the use of momenta allow us to accelerate the training process. The choice of discretization scheme will strongly affect the efficiency and robustness of the resulting training method. Hence, to allow for ease and efficacy of implementation of our methods, we describe the most appropriate discretization schemes in detail in Appendix \ref{sec:discretization} for both the general setting and for the specific constraints we consider in this paper, i.e., circle and orthogonality constraints. 

\subsection{Langevin Dynamics}\label{subsec:Langevin}
Consider the unconstrained Langevin system of SDEs
	\begin{align}
	\mathrm{d} \theta_t &= p_t \,\mathrm{d} t, \label{eq:UnderdampedLangevin_sde} 
	\\ \mathrm{d} p_t &= (-\nabla L(\theta_t) - \gamma p_t) \,\mathrm{d} t +  \sqrt{2\gamma\tau}\,\mathrm{d} \mathcal{W}_t, \nonumber
	\end{align}
with momenta $p$, parameters $\theta$, loss $L(\theta)$, temperature hyperparameter $\tau\geq 0$, friction hyperpar. $\gamma$, and $d$-dim. Wiener process $\mathcal{W}$ \cite{LeMa2015}. Under some mild assumptions, Langevin dynamics is provably ergodic, which means that its solutions sample the distribution: 
\begin{align}
    \rho \propto \ \text{exp}[-(L(\theta)+\|p\|^2/2)/\tau].
\end{align}
The temperature hyperparameter $\tau$, which controls the additive noise level, provides a direct connection between 
a pure optimization and sampling approach. The standard Bayes posterior is recovered for $\tau = 1$, whereas setting $\tau = 0$ will provide maximum a posteriori (MAP) point estimates. The range of values in between corresponds to an artificially sharpened posterior, where as $\tau \rightarrow 0$, the posterior probability mass is confined closer and closer to the modes of the distribution.\footnote{Techniques such as annealing or simulated tempering vary $\tau$ throughout training to enhance the optimization process \cite{annealing,ST}.} Using low temperatures \cite{LMV,coldposterior}, sampling methods have been found to enhance exploration and speed the approach to `good' minima, which enhance their generalization to nearby data sets. In this work we therefore consider a constrained SDE approach to neural network training to allow for the incorporation of both constraints and additive noise. 

%
\subsection{Constrained Langevin: Ergodicity and Central Limit Theorem.}\label{subsec:constrainedLangevin}
The neural network loss function naturally extends to the variable $q=(\theta,\xi)\in\mathbb{R}^d$ 
taking the form $V(q) = L(\theta)$ (note that in particular $\nabla_\xi V = 0$). 
The first continuous training method we consider is the constrained overdamped Langevin\footnote{Unconstrained stochastic gradient overdamped Langevin dynamics is analogous to the algorithm known as SGLD \cite{WeT11} in the machine learning literature. In SGLD one adds an additional additive noise term (typically with constant variance) to the dynamics. For a decaying sequence of stepsizes $h_n \rightarrow$ 0 one expects to eventually sample from a known stationary distribution.} system
\begin{align}
	\label{eq:constrainedOverdampedLangevin}
	\mathrm{d} q_t &= -\nabla V(q_t)\,\mathrm{d} t +  \sqrt{2\tau}\,\mathrm{d} \mathcal{W}_t  - \nabla_q g(q_t)\,\mathrm{d} \lambda_t, \\ 
    0 &= g(q_t), \nonumber
\end{align}
where $\mathcal{W}$ is a $d$-dim. Wiener process, $\tau\geq 0$ is the temperature hyperparameter,
and $\lambda_t$ is an $\mathbb{R}^m$-valued vector of Lagrange multipliers.
Provided the initial configuration $q_0$ satisfies the constraint, any trajectory $q_t$ of \eqref{eq:constrainedOverdampedLangevin} remains on the constraint manifold $\Sigma$ defined in Eq. \eqref{eq:constraintManifold}.
For $\beta^{-1}=\tau>0$, \eqref{eq:constrainedOverdampedLangevin} is equivalent to an underlying  ergodic (unconstrained) SDE (see Appx. \ref{sec:proofUnderlyingSDE}) with unique invariant measure 
\begin{equation}	\label{eq:OL_invariantMeasure}
\mathrm{d}\nu_\Sigma = Z^{-1} e^{-\beta V(q)} \, \mathrm{d}\sigma_\Sigma,
\ 
Z = \int_\Sigma e^{-\beta V(q)} \, \mathrm{d}\sigma_\Sigma,
\end{equation}
where $\sigma_\Sigma$ is the surface measure on $\Sigma$.

Ergodicity ensures that averages of observables with respect to $\nu_\Sigma$ can be approximated by time averages of trajectories of \eqref{eq:constrainedOverdampedLangevin}. 
To ensure the practical use of \eqref{eq:constrainedOverdampedLangevin} as a training method, we need the convergence to occur in a reasonable time.  
Thanks to the reversibility of the underlying SDE (see Appx. \ref{sec:proofUnderlyingSDE}), exponential convergence to equilibrium occurs as a consequence of a Poincar\'e inequality for $\nu_\Sigma$ (see Appx. \ref{sec:proofPoincare}, \ref{sec:proofCorollaries} and \citet{BGL13}). We provide a summary of the results here and refer to Appx. A for more details. 

A Poincar\'e inequality holds under a curvature-dimension assumption: there exists $\rho>0$ such that
\begin{equation} \label{eq:assumptionV}
C\!D(\rho,\infty):\qquad
\quad\mathrm{Ric}_\mathfrak{g} + \beta \nabla_\mathfrak{g}^2 V
 \geq \rho \mathfrak{g},
\end{equation}
in the sense of symmetric matrices.
The terms in \eqref{eq:assumptionV} rely on the structure of $\Sigma$ as a Riemannian manifold:
$\mathfrak{g}$ is the Riemannian metric, 
$\mathrm{Ric}_\mathfrak{g}$ is the Ricci curvature tensor and
$\nabla_\mathfrak{g}^2 V$ is the Hessian of $V$ on the manifold.
Under \eqref{eq:assumptionV} we have the following result (\cite{BGL13}, Appx. \ref{sec:proofPoincare}).
\vspace*{-0.2cm}
\begin{theorem}	\label{thm:OL_poincare}
Assume that there exists $\rho>0$ and $N>n$ such that $CD(\rho,N)$ holds.
Then $\nu_{\Sigma}$ satisfies a Poincaré inequality:
there exists a constant $L>0$ such that
\begin{align}	\label{eq:OL_poincare}
&\int\limits_{\Sigma} \big|\phi(q) -\intMean{\phi}_{\nu_\Sigma} \big|^2 \,\text{d}\nu_\Sigma(q) 
	\leq \frac{1}{2L} \int\limits_{\Sigma} \big|\Pi(q)\nabla \phi(q)\big|^2 \,\text{d}\nu_\Sigma \nonumber \\
&\forall \phi\in H^1(\nu_\Sigma),
\end{align}
where $\Pi(q)$ is the projection onto the cotangent space $T_q^*\Sigma$ Eq. (\ref{eq:OL_definitionPi}) and
$H^1(\nu_\Sigma)$ is the space of functions with square $\nu_\Sigma$-integrable gradients Eq. (\ref{eq:defH1}).
\end{theorem}
Consequences of Theorem \ref{thm:OL_poincare} are the exponential convergence and a central limit theorem (CLT) for the convergence in Eq. (\ref{eq:ergodicity}) (see Appx. \ref{sec:proofCorollaries}).
\vspace*{-0.2cm}
\begin{corollary}	\label{cor:geometricErgodicityAndClt}
If \eqref{eq:assumptionV} holds then
\begin{align}	\label{eq:geometricErgodicity}
&\int_\Sigma \big| \mathbb{E}( \phi(q_t) \mid q_0 ) - \intMean{\phi}_{\nu_\Sigma} \big|^2 \,\text{d}\nu_{\Sigma}(q_0)
	 \leq C(\phi)e^{-2L/\beta t}\ \nonumber\\
&\forall \phi\in H^1(\nu_\Sigma),
\end{align}
where $C(\phi)$ depends only on $\phi$.
Furthermore we have the following convergence in law:
\[
\sqrt{T}\big( \intMean{\phi}_T - \intMean{\phi}_{\nu_\Sigma} \big) \to \mathcal{N} ( 0 , \sigma_\phi^2 )
\quad \text{as }T\to\infty,
\]
where the asymptotic variance $\sigma_\phi^2$ is bounded as \\
$\sigma_\phi^2 \leq  \frac{\beta}{L} \int_{\Sigma} \big|\phi -\intMean{\phi}_{\nu_\Sigma} \big|^2 \,\text{d}\nu_\Sigma$.
\end{corollary}

In $\mathbb{R}^n$ assumption \eqref{eq:assumptionV} is equivalent to  convexity of $V$, which is known to be too strong a requirement (a confining assumption is sufficient, see e.g. \citet{LelS16}).
Although \eqref{eq:assumptionV} can certainly be weakened, the above results ensure that provided the curvature of the manifold is well behaved, sampling on $\Sigma$ has similar properties as on a flat space.%

Introducing momenta $p$ leads to constrained underdamped Langevin dynamics, 
the 2nd order counterpart of Eq. \eqref{eq:constrainedOverdampedLangevin}
	\begin{align}
	\mathrm{d} q_t &= p_t \,\mathrm{d} t,\ \ 0 = g(q_t),\ \label{eq:constrainedUnderdampedLangevin_sde} 
	\\ \mathrm{d} p_t &= (-\nabla V(q_t) - \gamma p_t) \,\mathrm{d} t +  \sqrt{2\gamma\tau}\,\mathrm{d} \mathcal{W}_t  - \nabla g(q_t)\mathrm{d} \lambda_t,\nonumber
	\end{align}
where $\gamma$ is the friction hyperparameter. 
The constraint 
induces a cotangency condition:
$p\in T_q^*\Sigma$, where $T_q^*\Sigma = \{ p\in\mathbb{R}^d \mid \nabla^T g(q) p = 0\}$ is the cotangent space of the manifold $\Sigma$. 
The corresponding phase space is the cotangent bundle 
$T^*\Sigma = \{ (q,p) \mid q\in\Sigma, p\in T_q^*\Sigma\}$.
Given an initial pair $(q,p) \in T^*\Sigma$, any trajectory $(q_t,p_t)$ of \eqref{eq:constrainedUnderdampedLangevin_sde}  stays on $T^*\Sigma$ for all time. 

\eqref{eq:constrainedUnderdampedLangevin_sde} is equivalent to an underlying ergodic SDE, whose invariant measure is 
$\mathrm{d}\mu = e^{-\beta H(q,p)} \mathrm{d}\sigma_{T^*\Sigma}$, with Hamiltonian $H(q,p) = V(q) + \frac{1}{2} p^Tp$ and
Liouville measure of the cotangent bundle $\sigma_{T^*\Sigma}$ \cite{LRS12}.
Based on the result for the unconstrained case, we expect exponential convergence to equilibrium also to hold here, but will leave this technical proof (e.g. based on hypocoercivity \cite{Vil09, LelS16}) for future work.

\subsection{Discretization of Constrained Langevin Dynamics.}\label{subsec:discretization}
The simplest iteration scheme $q_n\in\Sigma\mapsto q_{n+1}\in\Sigma$ for constrained overdamped Langevin dynamics \eqref{eq:constrainedOverdampedLangevin} consists of an Euler--Maruyama step followed by projection onto the constraint manifold $\Sigma$. 
The best choice for the projection is constraint-specific. 

For circle constraints we suggest orthogonal projection, which is both explicit and robust (we describe this in detail in Appx. \ref{sec:discr_OLcircle}).   
For orthogonality constraints, we derive an efficient quasi-Newton scheme to solve the non-linear system for the projection step (Appx. \ref{sec:OL_orthogonalConstraint}). We present the resulting training scheme in Algorithm \ref{eq:OL_OG_main}, where we denote $Q = W^\ell$ if $n^\ell\leq n^{\ell-1}$ and $Q = (W^\ell)^T$ otherwise, 
and present one training iteration $Q_n\in\Sigma\mapsto Q_{n+1}\in\Sigma$. Further, we denote $h$ as the stepsize, $G(Q) = \nabla_Q V(Q)$ and $\tilde{G}$ the gradient of the loss evaluated on a randomly subsampled partial data set. $R_n$ is an independent standard random normal matrix of the same size as $Q$. The initialization must be done with care: the constrained parameters and the potential slack variable must satisfy the constraint initially. 
\begin{algorithm}[h]
   \caption{Orthog. constraint overdamped Langevin}
   \label{eq:OL_OG_main}
\begin{algorithmic}
  \STATE \hspace*{-0.4cm}  {\bfseries Every step:} 
   \STATE $Q^{(0)} = Q_n  - h \tilde{G}(Q_n) + \sqrt{2\tau h} R_n$,
   \FOR{$k=0$ {\bfseries to} $K-1$}
   \STATE $Q^{(k+1)} = Q^{(k)} - \tfrac{1}{2}Q_n\big( (Q^{(k)})^T Q^{(k)} - I_s \big)$,\\ 
   \ENDFOR
    \STATE $Q_{n+1} = Q^{(K)}$.
\end{algorithmic}
\end{algorithm}

For underdamped Langevin dynamics a common way of building discretization schemes is via the use of splitting methods \cite{LeM16}. For the constrained underdamped Langevin system \eqref{eq:constrainedUnderdampedLangevin_sde} 
an ABO splitting strategy 
under $0=g(q_t),\ \  0 = \nabla_qg(q_t)p_t$ gives: 
\begin{align}
{\text{A:}} \ 
\mathrm{d} q_t &= p_t \,\mathrm{d} t,\quad
\mathrm{d} p_t = - \nabla_qg(q_t)\,\mathrm{d} \lambda_t, 
\nonumber \\ 
{\text{B:}} \ 
\mathrm{d} q_t &=0, \ 
\mathrm{d} p_t = -\nabla_q V(q_t)\,\mathrm{d} t -\nabla_q g(q_t)\,\mathrm{d} \mu_t, 
\label{eq:Bcomponent}\\
{\text{O:}}\ 
\mathrm{d} q_t &=0, \  \mathrm{d} p_t = -\gamma p_t\,\mathrm{d} t + \sqrt{2\gamma \tau}\,\mathrm{d} \mathcal{W}_t -\nabla g(q_t)\,\mathrm{d} \nu_t, \ 
\nonumber 
\end{align}
In the specific case $\tau=0$ and by re-scaling $\mu =e^{-\gamma h}/h $ and $\delta t = h^2$, an OBA sequence is equivalent to the standard PyTorch form of SGD with momentum $\mu$ and stepsize $\delta t$ \cite{Pytorch, LMV}. As alternative one could use a symmetric splitting method, e.g. BAOAB method \cite{LMS16}, but this would lose its accuracy order advantage in the presence of gradient noise. 

In \eqref{eq:Bcomponent} the B and O components can be solved exactly (in law) while the A component can be approximated using a standard scheme for constrained ODEs (e.g. SHAKE or RATTLE \cite{LeR04}[Chap. 7]). Importantly, the A component does not involve the evaluation of the gradient.
For circle constraints the A step can be solved explicitly and the corresponding algorithm is provided in detail in Appendix \ref{sec:discr_ULcircle}.
For orthogonality constraints all details are provided in Appendix \ref{sec:UL_orthogonalConstraint}, but we will provide the algorithm here. 
For $Q\in\Sigma$, the projection onto the cotangent space $T^*_Q\Sigma$ is defined as $\Pi_Q : \mathbb{R}^{r\times s} \to \mathbb{R}^{r\times s}$,
\begin{align}
\bar P \mapsto \Pi_Q\bar P = \bar P - \frac{1}{2} Q (\bar P^TQ + Q^T\bar P).\label{eq:projectmom}
\end{align}
We initialize the parameters and momenta (using projection \eqref{eq:projectmom}) to obey the constraint. Then the ABO steps $(Q_n,P_n)\in T^*\Sigma \mapsto (Q_{n+1},P_{n+1})\in T^*\Sigma$
are given by Algorithm \ref{eq:UL_OG_main}, where $\tilde{G}(Q)$ is the gradient of the loss evaluated on a subset of the data. 
More details in Appx. \ref{sec:discretization}.

\begin{algorithm}[h]
   \caption{Orthog. constraint underdamped Langevin}
   \label{eq:UL_OG_main}
\begin{algorithmic}
  \STATE \hspace*{-0.4cm}  {\bfseries Every step:} 
   \STATE $Q^{(0)} = Q_n  + hP_{n}$, \tikzmark{top} 
   \FOR{$k=0$ {\bfseries to} $K-1$}
   \STATE $Q^{(k+1)} = Q^{(k)} - \tfrac{1}{2}Q_n\big( (Q^{(k)})^T Q^{(k)} - I_s \big)$,\\ 
   \ENDFOR
    \STATE $Q_{n+1} = Q^{(K)},\  \bar P_{n+1} = P_n + \tfrac{1}{h}\big(Q_{n+1}-Q^{(0)}\big)$, \tikzmark{right} \\
     \STATE $P_{n+1} = \Pi_{Q_{n+1}} \bar P_{n+1}$ \tikzmark{bottom} \\ \tikzmark{topB}
     \STATE $\bar P_{n+1} = P_n - h \tilde{G}(Q_n)$, \\ 
     \STATE $P_{n+1} = \Pi_{Q_n}\bar{P}_{n+1}$,\tikzmark{bottomB} \\ \tikzmark{topC}
     \STATE $P_{n+1} = e^{-\gamma h} P_{n} + \sqrt{\tau(1-e^{-2\gamma h})}R_n$,\\
\STATE $P_{n+1} = \Pi_{Q_n} \bar P_{n+1}$ \tikzmark{bottomC}
\end{algorithmic}
     \AddNote{top}{bottom}{right}{(A)}
     \AddNote{topB}{bottomB}{right}{(B)}
     \AddNote{topC}{bottomC}{right}{(C)}
\end{algorithm}

\section{Numerical Experiments}\label{sec:numerics}
The use of constraints can enhance generalization performance. 
We support this claim by comparing the performance of neural network architectures trained using the constrained approaches described in this paper to nets trained using unconstrained SGD. We typically set $\tau=0$ and use equivalent learning rates to present a fair comparison between constrained and unconstrained approaches. We denote our circle and orthogonal Constrained \textit{overdamped} Langevin Algorithms as c-CoL\textit{od} and o-CoL\textit{od}, respectively. We compare \textit{underdamped} variants (CoL\textit{ud}) with SGD with momentum (SGD-m). 
\begin{figure*}[h]
    \begin{center}
    \centerline{\includegraphics[width = \textwidth]{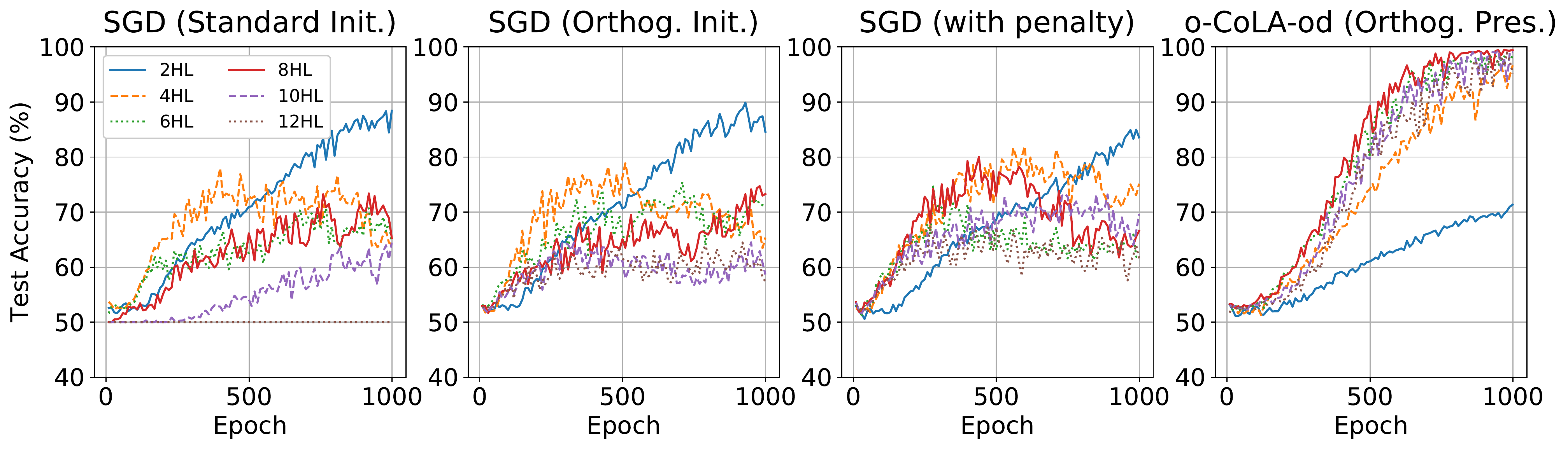}} 
    \vskip -0.2in
    \caption{Test acc. 
    of MLPs with $p$-number of 100-node hidden layers (HL), ReLU activation. 
    The MLPs are trained on a 4-turn spiral dataset (Fig. D\ref{Spiraldata}) 
    using SGD with standard initialization (1st column), SGD with orthogonal initialization (2nd col.), SGD with orthogonal regularization (`soft constraint') by adding a penalty term with strength $\lambda$ to the loss (3rd col.), and o-CoLod with $ \tau = 0$ (4th col.). For the orthogonal regularization approach and o-CoLod we constrain weights in all layers, apart from input and output layers. We set stepsize $h = 0.1$ for all methods and use 5\% subsampling. We found the optimal penalty strength $\lambda = 0.05$ for the orthogonal regularization method through line search. 
    Results are averaged over 10 runs. We observe that our o-CoLod method significantly outperforms unconstrained SGD and SGD with a soft constraint for MLPs with more than 3 hidden layers.}\label{MLP_OCSGD}
    \end{center}
    \vskip -0.3in
\end{figure*}
\subsection{Orthogonality Constraints}
In Fig. \ref{MLP_OCSGD} we want to train a multi-layer perceptron (MLP) with $p$ hidden layers on a tightly wound spiral binary classification problem (Fig. D\ref{Spiraldata}) and compare the performance of SGD with our orthogonality-preserving overdamped Langevin method o-CoLod. For SGD we show results for i) standard PyTorch initialization, ii) orthogonal initialization, and iii) orthogonal regularization (`soft constraint'), where a penalty term is added to the loss to encourage orthogonality of the NN weight matrices. Our o-CoLod method clearly outperforms all of these variants in terms of test accuracy 
for MLPs with more than 3 hidden layers. In Appx. \ref{sec:NumericsAppx} (Fig. D\ref{Spiral_withT}) we show that the use of a small temperature perturbation can speed up training even further and slightly increase the test accuracy. The performance of the soft constraint approach can be somewhat improved by lowering the stepsize, yet cannot match the performance of o-CoLod (see Figure \ref{Stiffness}). This illustrates the undesirable stiffness introduced into the system by using penalty-based regularization. The use of o-CoLod also removes the need to tune an additional parameter (the penalty strength). 

For a ResNet-34 architecture with BatchNorm and learning rate (LR) decay on CIFAR-10 \cite{cifar10} data our underdamped orthogonal constrained method, without weight decay (WD)  significantly outperforms SGD-m without WD (Fig. \ref{OG-OBA_Resnet}). The overdamped case is presented in the supplement, Fig. D\ref{CSGD_Resnet}. In future work we will explore the nuances of combining orthogonality constraints with BatchNorm, residual connections and LR decay. 

\subsection{Circle Constraints} 
\begin{figure*}
    \centering
      \includegraphics[width=0.9\linewidth]{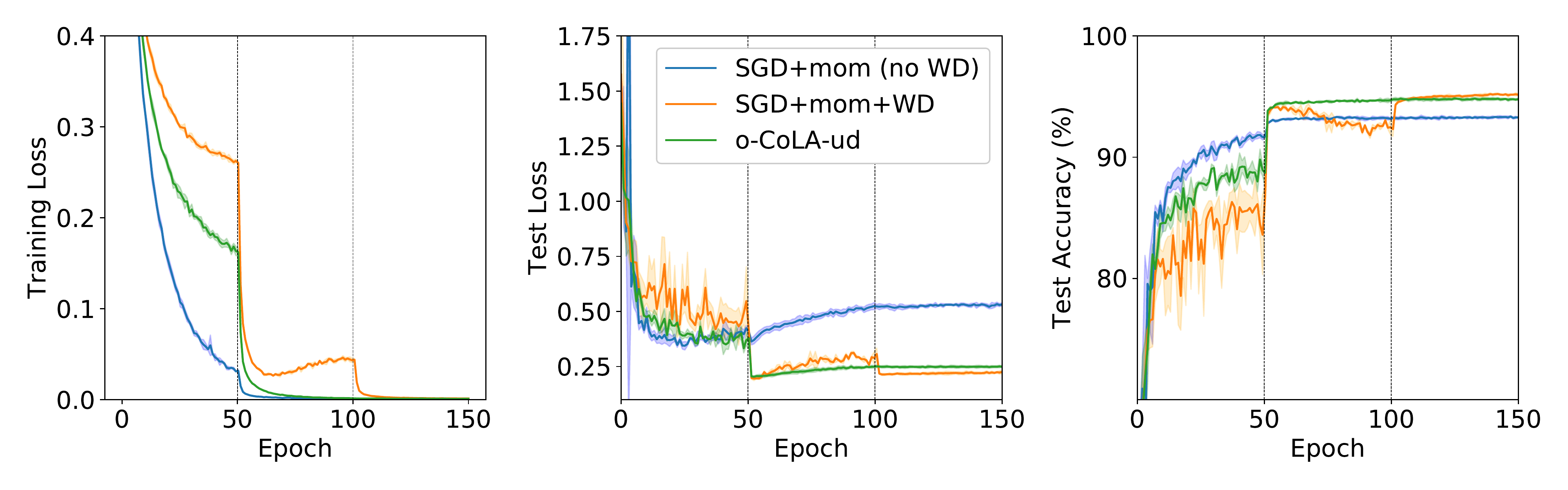} 
\vspace{-5mm}
    \caption{Train (left) \& test (middle) loss and test accuracy (right) averaged over 5 runs of a ResNet-34 with BatchNorm trained using SGD-m vs. o-CoLud with $\tau = 0$ on CIFAR-10. For SGD we initially use $h = 0.1$ and decay by a factor 10 every 50 epochs (indicated by the vertical black dotted lines). We set momentum = 0.9 and present results with and without WD. o-CoLud (with $\gamma = 0.5$) did not use WD. Its learning rate was re-scaled to match the parameters of SGD-m and used the same LR schedule. The o-CoLud method without weight decay strongly outperforms SGD-m without weight decay. \label{OG-OBA_Resnet}}
    \vspace*{7mm}
    \centering
      \includegraphics[clip, trim=0cm 6.5cm 0cm 6.5cm,width=\linewidth]{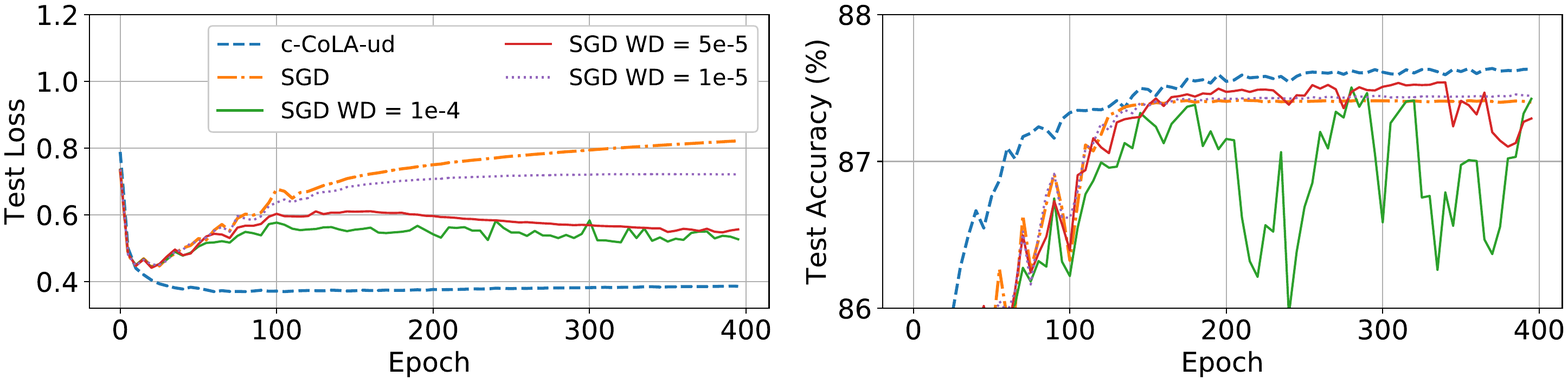} 
\vspace{-5mm}
    \caption{Test loss (left) and test accuracy (right) averaged over 5 runs of a 1000-node SHLP trained using SGD-m vs. c-CoLud on Fashion-MNIST (batchsize 128, number of training data samples reduced to 10K). After a line search we chose the best performing hyperparameter setting for SGD, namely $h = 0.1, mom = 0.8$, and varied the amount of weight decay (WD). Standard deviations are provided in the supplement. Hyperparameters c-CoLud: $h = 0.3, \gamma = 1, r_0 = 0.05, r_1 = 0.1, \tau = 0$. Due to the small training dataset size both methods quickly reached 100\% training accuracy, but c-CoLud is superior in its test loss and test accuracy. \label{CNet_FashionMNIST}}
\end{figure*}

We evaluate our circle constrained c-CoLud method on the Fashion-MNIST data set \cite{FashionMNIST}. 
We reduce the amount of training data to 10K samples and use the remaining 60K samples 
as test data. 
c-CoLud clearly outperforms SGD-m in terms of both test accuracy and test loss for a 1000-node single hidden layer perceptron (see Fig. \ref{CNet_FashionMNIST}). The lower test loss of c-CoLud is maintained during training and the method shows no signs of overfitting, thus eliminating the need for early stopping. 
Even with weight decay, SGD-m is outperformed by its constrained counterpart (for more detailed hyperparameter studies see Appx. \ref{sec:NumericsAppx}). We also show that a small transformer \cite{Transformer} with 2 encoder layers (each with 2-head self-attention and 200-node feed-forward network) trained using c-CoLud achieves a lower validation loss on NLP datasets than its unconstrained counterpart, SGD-m (Table \ref{NLPexp}).

 \begin{table}[!h]
 \vskip -0.1in
\caption{Minimum val. loss on Penn Treebank data (batchsize 1024) \cite{PennTreebank} and Wikitext-2 (batchsize 128) \cite{Wikitext2} using a transformer trained using c-CoLud or SGD-m. Hyperpar. c-CoLud: $h = 0.4, r = 0.5, r_L = 0.1, r_N = 1, r_A = 1, \tau = 0, \gamma = 0.5$ (Treebank) and $\gamma = 1$ (Wikitext-2), where the subscripts $L, N, A$ represent the radii belonging to the linear, norm and self- attention layers respectively. The transformer trained using c-CoLud obtains lower validation losses. Studies with weight decay are provided in the supplement.} \label{NLPexp}
\center
\begin{tabular}{c|c|c} 
      \textit{Optimizer} & Penn Treebank & Wikitext-2 
      \\ \hline 
      c-CoLud  & $\textbf{4.81}$
    & $\textbf{5.09}$ \\ \hline
  SGD $h$ = 0.1 \ \ \ \ \ \ \ \ \ \ \  & & \\
  \ \ \ \ \ \ \ \ \ $mom$ = 0.7 & 4.87 & 5.13  \\
  \ \ \ \ \ \ \ \ \ $mom$ = 0.8 & 4.83 & 5.13  \\
  \ \ \ \ \ \ \ \ \ $mom$ = 0.9 & 4.84 & 5.13  \\ \hline
  SGD $h$ = 0.2 \ \ \ \ \ \ \ \ \ \ \  & & \\
  \ \ \ \ \ \ \ \ \ $mom$ = 0.7 & 4.83 & 5.13  \\
  \ \ \ \ \ \ \ \ \ $mom$ = 0.8 & 4.83 & 5.14  
\end{tabular} 
\end{table}
\vskip -0.3in

\section{Conclusion} 
We provide a general framework that can be used to directly influence the parameter space of deep neural networks. 
The constrained SDE-based algorithms described in this paper allow for the use of additive noise to enhance exploration but can also be used directly in combination with standard SGD approaches. We provide a mathematical framework to study these regularized training methods as discretizations of constrained Langevin dynamics and provide detailed discretization schemes (see Appendix \ref{sec:discretization}). As specific examples of constraints we consider circle and orthogonality constraints, which obtain improved generalization performance on classification tasks compared to unconstrained SGD and soft constraint approaches. Further uses of our general framework are left for future work. 

\begin{figure*}[p]
    \centering
    \includegraphics[width=\textwidth]{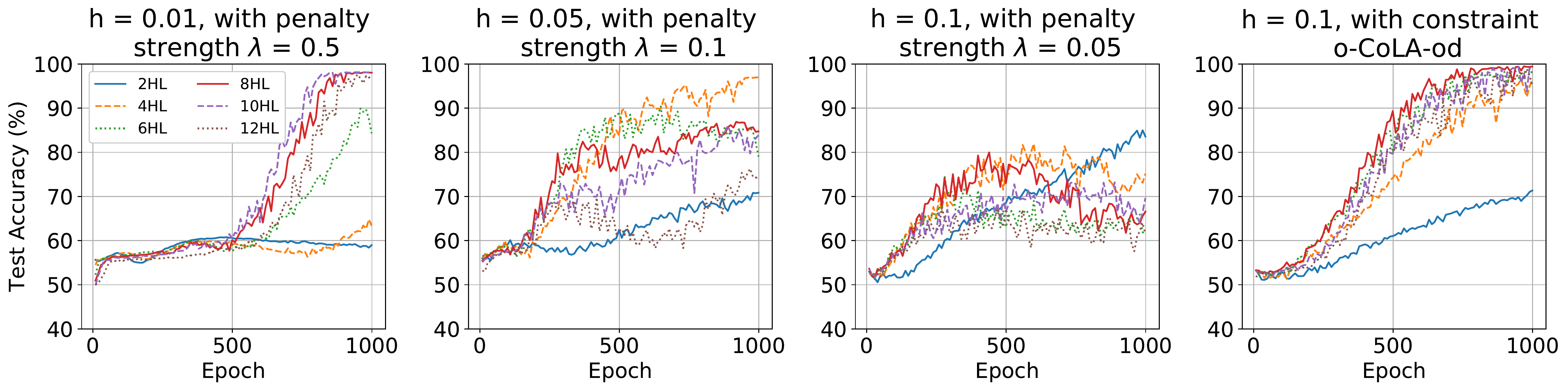}
     \vskip -0.15in
    \caption{Same set-up as for Figure \ref{MLP_OCSGD}. MLPs with varying numbers of hidden layers (HL) were trained using o-CoLod with $h = 0.1$ (right-most) and using SGD with a penalty term added to the loss (results are presented in the 1st three columns with varying stepsizes $h$ and penalty strengths $\lambda$). Results are averaged over 10 runs. We illustrate that the use of a penalty-based soft constraint introduces an undesirable stiffness into the system, needing the stepsize to be lowered to improve performance and to allow for the use of larger penalty strengths. The soft constraint approach is unable to reach the same performance as our o-CoLod method (right-most) and its performance is heavily dependent on the choice of penalty strength and step size.}\label{Stiffness} 
\end{figure*}

\section*{Acknowledgements}
The authors wish to thank Gabriel Stoltz and Tony Lelièvre for helpful discussions on constrained SDEs. Benedict Leimkuhler is a fellow of the Alan Turing Institute which is supported by EPSRC grant EP/N510129/1. During the creation of this paper Timothée Pouchon was supported by the Swiss National Science Foundation, project P2ELP2\_188037. Tiffany Vlaar is supported by The Maxwell Institute Graduate School in Analysis and its Applications, a Centre for Doctoral Training funded by the UK Engineering and Physical Sciences Research Council (grant EP/L016508/01), the Scottish Funding Council, Heriot-Watt University and the University of Edinburgh.

\nocite{BaE85}
\nocite{Bha82}
\nocite{Pytorchinit}
\nocite{KiV86}
\nocite{levelsetmethod}
\newpage
\newpage
\bibliography{references}
\bibliographystyle{icml2021}

\newpage
\appendix


\section{Theory of constrained overdamped Langevin dynamics}	\label{sec:supplementSDEs}

We present here the details of the theory summarized in Sec. \ref{sec:constrainedSDEs}.	
In particular, we provide the key results to establish the exponential convergence to equilibrium of
constrained overdamped Langevin dynamics Eq. (\ref{eq:constrainedOverdampedLangevin}).

In the first part (Sec. \ref{sec:proofUnderlyingSDE}), we derive the underlying SDE associated with Eq. (\ref{eq:constrainedOverdampedLangevin}), its generator and
the invariant measure $\nu_\manifold$ defined as
\begin{equation}	\label{eq:OL_invariantMeasureA}
\d\nu_\manifold = Z^{-1} e^{-\beta V(q)} \, \d\sigma_\manifold,
\quad
Z = \int_\manifold e^{-\beta V(q)} \, \d\sigma_\manifold,
\end{equation}
where $\sigma_\manifold$ is the surface measure on $\manifold$. Ergodicity ensures that averages of observables with respect to $\nu_\manifold$ can be approximated by time averages of trajectories of Eq. (\ref{eq:constrainedOverdampedLangevin}): for all test function $\phi\in\fcnC^\infty_c(\manifold)$ 
\begin{align}	\label{eq:ergodicity}
&\lim\limits_{T\to\infty}\intMean{\phi}_T = 
\intMean{\phi}_{\nu_{\manifold}}
\quad \text{for a.e. }q_0\in\manifold,  \nonumber\\
&\intMean{\phi}_T:= \frac{1}{T} \int_0^T \phi(q_t)  \,\d t
,\ 
\intMean{\phi}_{\nu_{\manifold}}:=\int_{\manifold} \phi(q) \,\d\nu_{\manifold}(q)
.
\end{align}

Next, in Sec. \ref{sec:proofPoincare} we present the Poincar\'e inequality on a manifold, which holds under a curvature-dimension assumption: there exists $\rho>0$ such that
\begin{equation} \label{eq:assumptionVA}
C\!D(\rho,\infty):\qquad
\quad\mathrm{Ric}_\frakg + \beta \nabla_\frakg^2 V 
 \geq \rho \frakg,
\end{equation}
in the sense of symmetric matrices, where
$\frakg$ is the Riemannian metric, 
$\mathrm{Ric}_\frakg$ is the Ricci curvature tensor and
$\nabla_\frakg^2 V$ is the Hessian of $V$ on the manifold.
Under Eq. \eqref{eq:assumptionVA} the following result holds. 
\begin{theorem}	\label{thm:OL_poincareA}
Assume that there exists $\rho>0$ and $N>n$ such that $CD(\rho,N)$ holds.
Then $\nu_\manifold$ satisfies a Poincaré inequality:
there exists a constant $L>0$ such that
\begin{align}	\label{eq:OL_poincareA}
&\int\limits_{\manifold} \big|\phi(q) -\intMean{\phi}_{\nu_\manifold} \big|^2 \,\d\nu_\manifold(q) \nonumber \\
& \ \ \ \ \ \ \ \ \ 	\leq \frac{1}{2L} \int\limits_{\manifold} \big|\Pi(q)\nabla \phi(q)\big|^2 \,\d\nu_\manifold(q) \nonumber \\
&\forall \phi\in H^1(\nu_\manifold),
\end{align}
where $\Pi(q)$ is the projection onto the cotangent space $T_q^*\manifold$ Eq. \eqref{eq:OL_definitionPi} and
$H^1(\nu_\manifold)$ is the space of functions with square $\nu_\manifold$-integrable gradients Eq. \eqref{eq:defH1}.
\end{theorem}
Consequences of Theorem \ref{thm:OL_poincareA} are the exponential convergence and a central limit theorem (CLT) for the convergence in Eq. \eqref{eq:ergodicity}

\begin{corollary}	\label{cor:geometricErgodicityAndCltA}
If Eq. \eqref{eq:assumptionVA} holds then
\begin{align}	\label{eq:geometricErgodicityA}
& \int_\manifold \big| \E( \phi(q_t) \mid q_0 ) - \intMean{\phi}_{\nu_\manifold} \big|^2 \,\d\nu_{\manifold}(q_0)
	 \leq C(\phi)e^{-2L/\beta t} \nonumber \\
& \forall \phi\in H^1(\nu_\manifold),
\end{align}
where $C(\phi)$ depends only on $\phi$.
Furthermore we have the following convergence in law:
\[
\sqrt{T}\big( \intMean{\phi}_T - \intMean{\phi}_{\nu_\manifold} \big) \to \mathcal{N} ( 0 , \sigma_\phi^2 )
\quad \text{as }T\to\infty,
\]
where the asymptotic variance $\sigma_\phi^2$ is bounded as \\
$\sigma_\phi^2 \leq  \frac{\beta}{L} \int_{\manifold} \big|\phi -\intMean{\phi}_{\nu_\manifold} \big|^2 \,\d\nu_\manifold$.
\end{corollary}
Appx. \ref{sec:proofCorollaries} is dedicated to using the Poincar\'e inequality to proving this. 

\subsubsection*{Notation}
We collect here additional notation needed for this discussion.

Given a measure $\mu$ in a space $E\subset \R^d$, we associate the space of square integrable functions
\begin{equation*}	
\Ld(\mu)= \big\{ \phi:E\to\R \text{ measurable } : \int_{E} |\phi|^2\,\d \mu < \infty \big\}.
\end{equation*}
Equipped with the inner product and associated norm
\[
\ps{ \phi, \psi }_{\mu} = \int_{E} \phi \psi\,\d \mu,
\qquad
\norm{\phi}_{\Ld(\mu)} = \sqrt{\psbig{ \phi, \phi }},
\]
$\Ld(\mu)$ is a Hilbert space.
We further define the subspace  $\Ld(\mu)$ of functions with zero mean by
\begin{equation}
\Ld_0(\mu) = \big\{ \phi\in \Ld(\mu) : \intMean{\phi}_\mu = 0\big\},
\quad
\intMean{\phi}_\mu = \int_{E} \phi \d \mu,
\end{equation}
as well as the space of functions with square integrable gradient
\begin{equation}	\label{eq:defH1}
H^1(\mu) = \big\{ \phi\in \Ld(\mu): \partial_i\phi\in \Ld(\mu) \quad 1\leq i\leq d\big\}.
\end{equation}

For the constraint $g:\R^d\to\R^m$,
we denote the Jacobian matrix as $G(q) = \nabla_q^Tg(q)$ and denote its right pseudo-inverse by $G^{+} = G^T(GG^T)^{-1}$ ($GG^T$ is invertible if $G$ has full row rank).
We verify that the map
\begin{equation}	\label{eq:OL_definitionPi}
\Pi:\R^d\to\R^{d \times d}, 
\quad q\mapsto \Pi(q) = I_d - G^+(q)G(q),
\end{equation}
defines for each $q$ the orthogonal projection onto the cotangent space $T_q^*\manifold$.
\[
\Pi_q=\Pi(q) : \R^d\to\R^d,
\quad p\mapsto \Pi(q)p .
\]
In particular, for all $q$ we have $\Pi_q p\in T_q^*\manifold$ and the matrix $\Pi_q$ is symmetric and idempotent:
(i.e., $\Pi_q^T=\Pi_q$ and $\Pi_q^2=\Pi_q$).

\subsection{The underlying SDE and the invariant measure}	\label{sec:proofUnderlyingSDE}

Although presented differently, the results of this section follow closely the treatment of this issue presented in \cite{LSR10}.

We define the mean curvature of the manifold as the vector valued function
\begin{align}	\label{eq:OL_definitionH}
\mathcal{H}:\R^d\to\R^d,
\quad q\mapsto  \big(\mathcal{H}(q)\big)_i = \Pi_{jk}(q)\partial_j \Pi_{ik}(q) \\
1\leq i\leq d, \nonumber
\end{align}
where $\Pi(q):\R^d\to\R^d$ is the projection onto the cotangent space defined in Eq. \eqref{eq:OL_definitionPi}.
We then establish the following result (proved below).
\begin{lemma}	\label{lem:OL_augmentedDynamics}
	The constrained system Eq. (\ref{eq:constrainedOverdampedLangevin}) can be rewritten as the following SDE in $\R^d$
	\begin{align}	\label{eq:OL_augmentedDynamics}
	\d q_t &= -\Pi(q_t) \nabla V(q_t)\d t +  \sqrt{2\beta^{-1}}\,\Pi(q_t)\d \mathcal{W}_t \nonumber \\
	&+ \beta^{-1}\mathcal{H}(q_t) \,\d t.
	\end{align}
\end{lemma}


The uniqueness of the invariant measure of Eq. \eqref{eq:OL_augmentedDynamics} and the resulting ergodicity result Eq. \eqref{eq:ergodicity} are proved in \cite{LSR10}[Prop. 3.20] 
(the proof relies on the divergence theorem on manifolds).

The generator associated with Eq. \eqref{eq:OL_augmentedDynamics} is given by
\[
\calL = -\Pi(q) \nabla V(q)\cdot \nabla + \beta^{-1}\mathcal{H}(q) \cdot \nabla + \beta^{-1}\Pi(q):\nabla^2.
\]
We verify that $\calL$ can be written in the following symmetric form 
\begin{align}	\label{eq:OL_Lsymmetric}
\calL\psi 
&=	\beta^{-1}\Div_\manifold(\nabla_\manifold\psi) - \nabla_\manifold V(q) \cdot \nabla_\manifold\psi \nonumber \\
&=	\beta^{-1} e^{\beta V(q)} \Div_\manifold\big( e^{-\beta V(q)} \nabla_\manifold \psi\big),
\end{align}
where we denote
$\nabla_\manifold \phi = \Pi\nabla\phi$
and 
$\Div_\manifold \psi = \nabla_\manifold\cdot\psi = \sum_{i,j=1}^d \Pi_{ij}\partial_j \psi_i$.
This expression directly implies that $\calL$ is reversible with respect to $\nu_{\manifold}$:
\begin{equation}	\label{eq:OL_reversibility}
\psbig{ \calL\phi, \psi }_{\nu_\manifold} = -\beta^{-1} \psbig{ \nabla_\manifold \phi, \nabla_\manifold \psi }_{\nu_\manifold}
= \psbig{ \phi, \calL\psi }_{\nu_\manifold}.
\end{equation}


Thanks to this expression, we can prove that the measure $\nu_{\manifold}$ is indeed invariant for Eq.(\ref{eq:constrainedOverdampedLangevin}).
Let us introduce the forward Kolmogorov equation: given a test function $\phi\in\fcnC^\infty_c(\manifold)$
\[
\partial_t u(t,q) = \calL u(t,q) ~t\geq0,~q\in\manifold\qquad u(0,q) = \phi(q).
\]
The solution to this equation is verified to be
$u(t,q) = \E(\phi(q_t)\mid q_0=q)$ (see the Feynmann--Kac formula)
and is usually denoted as $u(t,q) = e^{t\calL}\phi(q)$.
The measure $\nu_\manifold$ is invariant if for any $t\geq 0$
$\int_\manifold u(t,q) \,\d \nu_\manifold(q) = \int_\manifold u(0,q) \,\d \nu_\manifold(q) = \intMean{\phi}_{\nu_\manifold}$.
This is easily verified thanks to Eq. \eqref{eq:OL_reversibility}:
\begin{align*}
\frac{\d}{\d t} \int_\manifold u(t,q) \,\d \nu_\manifold(q)
=
\frac{\d}{\d t} \int_{\manifold} e^{t\calL}\phi(q) \,\d \nu_{\manifold}(q) \\
= \int_{\manifold} \calL e^{t\calL}\phi(q) \,\d \nu_{\manifold}(q)
= \psbig{ \calL e^{t\calL}\phi, \mathbf{1} }_{\nu_{\manifold}} = 0.
\end{align*}

\begin{proof} 
Let us write $\lambda_t$ as the It\^o process
\begin{equation}	\label{eq:lambdatAnsatz}
\d\lambda_t = \mu(q_t) \, \d t + \sigma(q_t) \, \d \mathcal{W}_t,
\end{equation}
where 
$\mu:\R^d\to\R^m$, $\sigma:\R^d\to\R^{m\times d}$
and $\mathcal{W}_t$ is the same Wiener process as in Eq. (\ref{eq:constrainedOverdampedLangevin}).
Using this expression in Eq. (\ref{eq:constrainedOverdampedLangevin}) brings
\begin{align*}
\d q_t = &\big(-\nabla V(q_t)-G(q_t)^T\mu(q_t)\big)\,\d t \\ 
&+ \big(\sqrt{2\beta^{-1}}I- G(q_t)^T\sigma(q_t)\big)\,\d \mathcal{W}_t,
\end{align*}
where we recall the notation for the Jacobian $G = \nabla_q^T g$.
Using It\^o formula we find 
\begin{align}	\label{eq:dgqt}
0&=\d g(q_t) = G(q_t)\,\d q + b_t\,\d t \nonumber \\
&= G(q_t)\big( -\nabla V(q_t) \,\d t+  \sqrt{2\beta^{-1}}\,\d \mathcal{W}_t  -G(q_t) ^T \d \lambda_t\big) \nonumber \\ 
& \ \ \ + b_t\,\d t,
\end{align}
where $b_t$ is the $d$-dimensional process defined as (omitting the dependence on $q_t$)
\begin{equation}	\label{eq:defbt}
\begin{aligned}
(b_t)_i &= \frac12 \big(\sqrt{2\beta^{-1}}I- G^T\sigma\big)\big(\sqrt{2\beta^{-1}}I- G^T\sigma\big)^T : \nabla^2 g_i
\\&= \beta^{-1}\Delta g_i
- \frac{\sqrt{2\beta^{-1}}}{2} \big(G^T\sigma+ \sigma^TG\big) :\nabla^2 g_i \\
&\ \ \ + \frac12 G^T\sigma\sigma^T G :\nabla^2 g_i.
\end{aligned}
\end{equation}
From Eq. \eqref{eq:dgqt} yields
\begin{align}	\label{eq:expressionLambda}
\d \lambda_t = & \ \big(G(q_t)G(q_t)^T\big)^{-1} G(q_t) \Big( -\nabla V(q_t) \,\d t \nonumber \\
&+ \sqrt{2\beta^{-1}} \,\d \mathcal{W}_t \Big) + \big(G(q_t)G(q_t)^T\big)^{-1} b_t \,\d t.
\end{align}
Identifying with Eq. \eqref{eq:lambdatAnsatz} we find \\ $\sigma(q) = \sqrt{2\beta^{-1}} (G^+(q))^T$, which used in Eq. \eqref{eq:defbt} yields
\begin{align*}
(b_t)_i = &\beta^{-1} \big(\Delta g_i - \big(G^T(G^+)^T+ G^+G\big) :\nabla^2 g_i \\
&+ G^T(G^+)^T G^+ G :\nabla^2 g_i\big ).
\end{align*}
As $G^+G$ is symmetric and $GG^+= I_m$, we obtain
\begin{equation}	\label{eq:expressionbt}
(b_t)_i = \beta^{-1}\big(\Delta g_i - G^+G :\nabla^2 g_i\big) = \beta^{-1} \Pi : \nabla^2 g_i.
\end{equation}
Inserting Eq. \eqref{eq:expressionLambda} in Eq. (\ref{eq:constrainedOverdampedLangevin}) brings
\begin{align}	\label{eq:expressiondqt}
\d q_t = &\ -\Pi(q_t) \nabla V(q_t)\d t +  \sqrt{2\beta^{-1}}\,\Pi(q_t)\d \mathcal{W}_t  \nonumber \\
&- G^+(q_t) b_t \,\d t.
\end{align}
To conclude the proof we require the following technical relations on the mean curvature vector
(Eq. \eqref{eq:expressionHwithdivPi} follows from a direct computation; the proof of Eq. \eqref{eq:equalityPiH} is direct but involved and can be found in \cite{LSR10}) 
\begin{lemma}	\label{lem:relationsCurvature}
	The projection $\Pi$ and the vector $H$ defined in Eq. \eqref{eq:OL_definitionPi} and Eq. \eqref{eq:OL_definitionH} satisfy the following equalities
	\begin{subequations}
		\begin{flalign}	
		\mathcal{H} &= (I-\Pi)\nabla\cdot \Pi,	
		\label{eq:expressionHwithdivPi}
		\\
		\Pi:\nabla^2g_i &= -(G\mathcal{H})_i\qquad	1\leq i\leq d,	
		\label{eq:equalityPiH}
		\end{flalign}
	\end{subequations}
\end{lemma}
Equality Eq. \eqref{eq:expressionHwithdivPi} ensures that $\Pi \mathcal{H} = 0$.
Combining Eq. \eqref{eq:expressionbt} and Eq. \eqref{eq:equalityPiH} we can write $b_t = -\beta^{-1}G\mathcal{H}$.
Thanks to these relations and the definition of $\Pi$, we obtain
\[
-G^+b_t = \beta^{-1}G^+G\mathcal{H} =\beta^{-1}(I-\Pi)\mathcal{H} = \beta^{-1} \mathcal{H}.
\]
This equality combined with Eq. \eqref{eq:expressiondqt} proves Eq. \eqref{lem:OL_augmentedDynamics} and concludes the proof of Lemma \ref{lem:OL_augmentedDynamics}. \\ 
\end{proof}

\subsection{Poincar\'e inequality on a manifold}	\label{sec:proofPoincare}

Poincar\'e inequalities, also called spectral gap inequalities, form an important family of functional inequalities in the theory of Markov diffusion processes.
They are the simplest inequalities that provide results on the convergence to equilibrium.
Stronger results can be obtained with the family of log-Sobolev inequalities, which are at the center of the Bakry--\'Emery theory \cite{BaE85}.
We follow here closely the book \cite{BGL13} on this subject (more specifically §1.16.2 and sections 4.2, 4.8, C.6).
For the necessary terminology of  Riemannian manifolds we recommend the introductory textbook \cite{Lee18} (the literature on this topic is vast and contains many works of high quality).

As presented in \cite{BGL13}[Chap. 4], 
a Poincar\'e inequality can be obtained as a consequence of a curvature-dimension condition.
For the sake of  presentation, we introduce this result in the setting of a weighted Riemannian manifold.
Let $(\calM,\frakg)$ be an $n$-dimensional Riemannian manifold, where $\frakg$ is the Riemannian metric.
We consider the diffusion operator
\[
\calL = \Delta_\frakg - \langle \nabla_\frakg W , \nabla_\frakg \cdot \rangle_\frakg ,
\]
where $\Delta_\frakg$ denotes the Laplace--Beltrami operator on the manifold $\calM$, $\nabla_\frakg$ denotes the Levi--Civita connection (covariant derivative) and $\langle\cdot,\cdot\rangle_\frakg$ denotes the Riemannian metric
($\langle X,Y\rangle_\frakg = \frakg(X,Y)$ for all vector fields $X,Y$).
We verify that the associated invariant measure is
$\d\mu = Z^{-1}e^{-W}\d\mu_\frakg$, where $\d\mu_\frakg$ is the Riemannian measure \cite{BGL13}[§1.11.3]. 
For $N\in[n,\infty]$, we define the 2-tensor 
\[
\mathrm{Ric}_{N}(\calL) = \mathrm{Ric}_\frakg + \nabla_\frakg^2 W - \frac{1}{N-n} \d W\otimes \d W.
\]
where $\mathrm{Ric}_\frakg$ is the Ricci curvature $2$-tensor and
$\nabla_\frakg^2$ denotes the Hessian operator on $\calM$ (the case $N=n$ is considered only if $W$ is constant).
In this context, a curvature-dimension condition $C\!D(\rho,N)$ for $\rho\in\R$ and $N\geq n$ holds if and only if (see \cite{BGL13}[C.6]) 
\begin{equation}	\label{eq:CD}
C\!D(\rho,N):\qquad\mathrm{Ric}_{N}(\calL) \geq \rho \frakg,
\end{equation}
in the sense of symmetric $(0,2)$-tensors (covariant 2-tensors).
In the flat space $\calM = \R^n$, the condition $C\!D(\rho,\infty)$ reads $\nabla^2 W\geq \rho I$, which is nothing but the convexity of the potential $W$.
Under $C\!D(\rho,N)$, the measure $\mu$ is proved to satisfy a Poincar\'e inequality 
(in \cite{BGL13}, combine Thm 4.8.4 with the discussion in section C.6).
\begin{theorem} \label{thm:poincareManifold}
\cite{BGL13}[Thm 4.8.4] 
Under the curvature-dimension condition $C\!D(\rho,N)$ with $\rho>0$ and $N\geq n$, $N>1$, 
the measure $\mu$ satisfies the Poincar\'e inequality
\begin{equation}	\label{eq:OL_poincare_app2}
\mathrm{Var}_\mu(\phi) = \norm{\phi-\intMean{\phi}_\mu}_{\Ld(\mu)}^2
	\leq C_P\norm{ \nabla_\frakg\phi }_{\Ld(\mu)}^2
\end{equation}
with constant $C_P = \frac{N-1}{\rho N}$, $\forall \phi\in \Ld(\mu) \cap H^1(\mu)$.
\end{theorem}

As the tensor $\d W\otimes \d W$ is positive semi-definite,
we verify the monotonicity $\mathrm{Ric}_{N+M}(\calL)\geq \mathrm{Ric}_{N}(\calL)$ for any $M\geq 0$.
This implies in particular that $CD(\rho,N)\Rightarrow CD(\rho,\infty)$ for any $N\in[n,\infty]$.
Hence, among all choices of $N\geq n$, $CD(\rho,\infty)$ is the weaker condition.

Let us now consider this result in the context of the constraint manifold $\manifold$ in Eq. (\ref{eq:constraintManifold}).
We consider the space $\R^d$ with its Riemannian manifold structure given by the Euclidean metric 
$\bar\frakg (v,w)= v\cdot w$ for all $v,w\in\R^d$ (for all $q\in\R^d$, $p\in T_q\R^d$ is identified with $\R^d$ through a canonical isomorphism).
Assuming that $g$ is smooth and that $\nabla_q^Tg$ has everywhere full row-rank, $\manifold$ is a smooth embedded submanifold of $\R^d$ of dimension $n = d-m$ (see e.g. \cite{Lee18}). 
Furthermore, $\manifold$ is equipped with the metric induced by $\bar\frakg$:
for a local parameterization of $\psi:U\subset\manifold\to\R^d$, $\bar \frakg$ is given locally on $U$ by 
\begin{equation}	\label{eq:inducedMetric}
\bar\frakg 
	= \sum_{i=1}^d\sum_{j,k=1}^n 
		\frac{\partial \psi^i}{\partial {x^j}} \frac{\partial \psi^i}{\partial {x^k}}  \,\d x^j \d x^k
	= \big(\nabla_x \psi \nabla_x^T \psi\big)_{jk} \,\d x^j \d x^k
\end{equation}
We now define the potential $W = \beta V|_\manifold$, where $V|_\manifold$ denotes the restriction of $V$ to $\manifold$.
Assumption \ref{eq:assumptionVA} corresponds then to condition $C\!D(\rho,\infty)$ above.
Applying Theorem \ref{thm:poincareManifold} we obtain Poincar\'e's inequality on the constraint manifold $\manifold$.
We note that for a function $\phi$ defined on $\R^d$, the covariant derivative in $\R^d$ of $\phi|_{\manifold}$ on the manifold is the orthogonal projection of the directional derivative of $\phi$ (in the ambient manifold $\R^d$) onto the cotangent space: 
$\nabla_\frakg(\phi|_{\manifold})(q) = \Pi(q)\nabla_q \phi(q)$.
Furthermore, we note that the surface measure $\sigma_\manifold$ equals the Riemannian measure on the manifold
(compare \cite{LSR10}[Rem. 3.4] 
with \cite{Lee18}[Prop. 2.41] 
and Eq. \eqref{eq:inducedMetric}).
We thus obtain the result of Theorem \ref{thm:OL_poincareA}
with constant $C_P = \frac1\rho = \frac{1}{2L}$.

\subsection{Exponential convergence to equilibrium and central limit theorem}	\label{sec:proofCorollaries}

Let us define the norm of a linear operator $\calA:\Ldznu\to\Ldznu$ as
\[
\norm{\calA}_{\mathcal{B}(\Ldznu)} = \sup_{\phi\in\Ldznu} \frac{\norm{\calA\phi}_{\Ldznu}}{\norm{\phi}_{\Ldznu}}.
\]
Denote $\bar{\phi} = \phi-\intMean{\phi}_{\nu_\manifold}\in\Ldznu$. The Poincar\'e inequality Eq. \eqref{eq:OL_poincareA}, rewritten on the subspace $\Ldznu$, is as follows:
\begin{equation}	\label{eq:OL_poincare_app}
\norm{\bar\phi}_{\Ldznu}^2	\leq \frac{1}{2L}\norm{ \nabla_\manifold\bar\phi }_{\Ldznu}^2
\  \forall\bar\phi\in \Ldznu \cap H^1(\nu_\manifold).
\end{equation}
Using the reversibility of the measure Eq. \eqref{eq:OL_reversibility}, we can prove the following result (the proof follows the same lines as \cite{LelS16}[Prop. 2.3], see also \cite{BGL13}[Thm 4.2.5]).
\begin{lemma} \label{lem:exponentialConvergence}
The measure $\nu_\manifold$ satisfies the Poincar\'e inequality Eq. \eqref{eq:OL_poincare_app} if and only if
\begin{equation}	\label{eq:exponentialConvergence}
\norm{e^{t\calL}}_{\mathcal{B}(\Ldznu)}
	\leq e^{-2\frac{L}{\beta}t}.
\end{equation}
\end{lemma}
Exponential convergence to equilibrium is then directly obtained from Lemma \ref{lem:exponentialConvergence}: 
\begin{align}	\label{eq:exponentialConv_app}
\norm{e^{t\calL}\bar\phi}_{\Ldznu}
	&\leq \norm{e^{t\calL}}_{\mathcal{B}(\Ldznu)} \norm{\bar\phi}_{\Ldznu}  \nonumber \\
	&\leq e^{-2\frac{L}{\beta}t} \norm{\bar\phi}_{\Ldznu}.
\end{align}
This inequality implies Eq. \eqref{eq:geometricErgodicityA} (note that $e^{t\calL} \intMean{\phi}_{\nu_\manifold} = \intMean{\phi}_{\nu_\manifold}$) and thus proves the first assertion of Corollary \ref{cor:geometricErgodicityAndCltA}.

A consequence of the exponential convergence to equilibrium Eq. \eqref{eq:exponentialConv_app} is the following central limit theorem for time averages $\intMean{\phi}_T = \frac{1}{T}\int_{0}^T \phi(q_t) \,\d t$ (see also \cite{KiV86}).
\begin{theorem}	\label{thm:clt_app}
	\cite{Bha82}
	If Eq. \eqref{eq:exponentialConv_app} holds, then the following convergence in law is satisfied
	\[
	\sqrt{T}\big( \intMean{\phi}_T - \intMean{\phi}_{\nu_\manifold} \big) \to \mathcal{N} ( 0 , \sigma_\phi^2 )
	\quad \text{as }T\to\infty,
	\]
	where the asymptotic variance $\sigma_\phi^2$ is given by the formula
	$\sigma_\phi^2	= 2 \ps{\bar\phi, -\calL^{-1} \bar\phi}$
	with $\bar\phi = \phi-\intMean{\phi}_{\nu_\manifold}$.
\end{theorem}

To quantify the asymptotic variance, we use the following classical result.
\begin{lemma}	\label{lem:exponentialConvergence_App2}
	(e.g., \cite{LelS16}[Prop. 2.1])
If Eq. \eqref{eq:exponentialConvergence} holds, then the generator $\calL$ is invertible and the resolvent can be expressed as $-\calL^{-1} = \int_{0}^\infty e^{t\calL} \,\d t$ and satisfies the bound
$\norm{\calL^{-1}}_{\mathcal{B}(\Ldznu)} \leq \frac{\beta}{2L}$.
\end{lemma}

Using Lemma \ref{lem:exponentialConvergence_App2} and Cauchy--Schwartz inequality, the asymptotic variance in Theorem \ref{thm:clt_app} can thus be bounded as
\begin{align*}
\sigma_\phi^2
	= 2 \int_{\manifold} \bar\phi (-\calL^{-1} \bar\phi) \,\d \nu_\manifold
	&\leq 2 \norm{\calL^{-1}}_{\mathcal{B}(\Ldznu)} \norm{\bar\phi}_{\Ldznu}^2 \\
	&\leq \frac{\beta}{L} \norm{\bar\phi}_{\Ldznu}^2.
\end{align*}
This estimate completes the proof of the second assertion of Corollary \ref{cor:geometricErgodicityAndCltA}.

\section{Discretization of constrained Langevin dynamics}	\label{sec:discretization}

We present here the details of the  constrained training methods considered in this paper.
Both the overdamped Eq. (\ref{eq:constrainedOverdampedLangevin}) and underdamped Eq. (\ref{eq:constrainedUnderdampedLangevin_sde}) Langevin dynamics are discretized for the constraints presented in Section \ref{sec:constraints}.
We emphasize that the initialization of each given method must be done with care: the constrained parameters, the potential slack variable, as well as their momenta in the underdamped case, have to satisfy the constraint initially.

Recall the notation introduced in Section \ref{sec:constraints}:
$\theta\in\R^{n}$ is the vector of all the parameters of the model,
we consider the variable $q = (\theta,\xi)\in\R^{d}$, $d=n+n^\xi$,
where $\xi\in\R^{s}$ is a slack variable to enforce the potential inequality constraints.
The loss is extended $q=(\theta,\xi)$ as
$V(q) = \loss(\theta)$ (in particular $\nabla_\xi V = 0$) 
and constraints are given by a map $g:\R^d\to\R^m$.
The parameters are partitioned as $\theta = (\theta^u,\theta^c)$,
where $\theta^u\in\R^{n^u}$ are not involved in any constraint
while $\theta^c\in\R^{n^c}$ are.

\subsection{Discretization of constrained overdamped Langevin (general constraint)}\label{sec:generalCoLAod}

Following \cite{LSR10}[Chap. 3] 
a simple discretization of the constrained overdamped Langevin dynamics Eq. (\ref{eq:constrainedOverdampedLangevin}) is given by the iteration
$q_n\in\manifold\mapsto q_{n+1}$ defined as
\begin{equation}
\label{eq:OLdiscretization1}
\begin{aligned}
&\bar{q}_{n+1} = q_n -\nabla_q V(q_n)h +  \sqrt{2\beta^{-1}h}\,R_n,  \\
&q_{n+1}= \bar{q}_{n+1}-\nabla_q g(q_n) \lambda_n,	\\
&\text{where $\lambda_n\in\R^m$ is such that }g(q_{n+1})=0,
\end{aligned}
\end{equation}
where $R_n\sim N(0,I)$ is a vector of iid standard normal random variable.
The first step of Eq. \eqref{eq:OLdiscretization1}, $\bar{q}_{n+1}$,
is an Euler--Maruyama step for standard overdamped Langevin.
As $\bar{q}_{n+1}$ in $\R^d$ is generally not on the constrained manifold $\manifold$, the last term is present to project $\bar{q}_{n+1}$ back onto  $\manifold$, ensuring  $g(q_{n+1})=0$.
In particular, for the unconstrained parameter we have 
$\nabla_{\theta^u}^Tg ={0}_{m\times n^u}$ which implies that $\theta^u_{n+1} = \bar\theta^u_{n+1}$ is a standard EM step.

In general, projecting back onto the manifold $\manifold$, i.e., finding $\lambda_n$,  can be done using root-finding algorithms.
Nevertheless, for certain constraints $g$ the roots can be found explicitly.
This is the case for the circle constraint Eq. (\ref{eq:circleConstraint})
(see Section \ref{sec:discr_OLcircle}). 
A potential weakness of method Eq. \eqref{eq:OLdiscretization1} is that the projection process can be guaranteed only for small enough step size $h$ (i.e. $\bar q_n$ must be close to $\manifold$).
Indeed, even for the circle constraint if $h$ is too large it might not be possible to project $\bar{q}_{n+1}$ back onto the circle following the direction $\nabla_q g(q_n)$.  See \cite{LSZ2020} for some discussion of methods to allow computation to be performed in the large timestep regime.

An alternative method is given by the iteration
$q_n\in\manifold\mapsto q_{n+1}\in\manifold$ defined as in \cite{LSR10}[Chap. 3]
\begin{equation}
\label{eq:OLdiscretization2}
\begin{aligned}
&\bar{q}_{n+1} = q_n -\nabla_q V(q_n)\d t +  \sqrt{2\beta^{-1}h}\,R_n,  \\
&q_{n+1} = \bar{q}_{n+1}-\nabla_q g({q}_{n+1}) \lambda_n,	\\
&\text{where $\lambda_n\in\R^m$ is such that }g(q_{n+1})=0,
\end{aligned}
\end{equation}
where $R_n\sim N(0,I)$ is a vector of iid standard normal random variable.
The projection used in method Eq. \eqref{eq:OLdiscretization2} is in general more robust.
The circle constraint is a good illustration of this: while in Eq. \eqref{eq:OLdiscretization1} we project following an oblique direction, in Eq. \eqref{eq:OLdiscretization2} the projection is orthogonal and always exists (see Section \ref{sec:discr_OLcircle}).

\subsection{Discretization of constrained underdamped Langevin (general constraint)}\label{sec:generalCoLAud}

We next consider  the discretization of the constrained underdamped Langevin dynamics Eq. (\ref{eq:constrainedUnderdampedLangevin_sde}) where we denote by $p = (p^u,p^c,p^\xi)\in\R^{n^u+n^c+n^\xi}$ the momenta associated with the configuration $q = (\theta^u,\theta^c,\xi)$.
Following \cite{LeM16}, the system is split into A,B,O components Eq. (\ref{eq:Bcomponent}), 
where B represents a projected impulse defined by the loss gradient (restricted to the cotangent space), O represents a projected stochastic impulse, and A represents evolution along geodesics (i.e., for circle constraints, these are rotations on the circles).

As in the overdamped case, the equality $\nabla_{\theta^u}^Tg = 0_{m\times n^u}$ ensures that the unconstrained parameters and their momenta $(\theta^u,p^u)$ evolve following the A,B,O steps for unconstrained underdamped Langevin (see \cite{LMS16}).
As the B and O components only involve a variation in the momentum $p_t$ and because the constraint only involves $q_t$, they can be solved exactly for any constraint. 
The A component involves a variation of the configuration $q_t$ and thus cannot be solved exactly (in law) for any constraint.
However, as this part does not include any force evaluation (which would require back-propagation to compute the gradient), it can be approximated cheaply using a few steps of standard well-known schemes such as SHAKE or RATTLE (see Section \ref{sec:UL_orthogonalConstraint} for orthogonal constraints).
Furthermore, for simple constraints such as the circle constraint Eq. (\ref{eq:circleConstraint}) the A component can be solved explicitly (see Section \ref{sec:discr_ULcircle}).

\def\confq{w}%
Let us present the details of the B and O steps.
For convenience, let us introduce the following notation for the variables involved in the constraint
$\confq=(\theta^c,\xi)\in\R^{n^c+n^\xi}$
and associated momentum 
$p^\confq=(p^c,p^\xi)\in\R^{n^c+n^\xi}$.
The projection onto the cotangent space Eq. \eqref{eq:OL_definitionPi} is then as 
\begin{align}	\label{eq:expressionPiw}
&\Pi(q) = I_d - 
\begin{pmatrix} 
0& 0 \\
0& \Pi_\confq(q)
\end{pmatrix},
\\ &\text{with }
\Pi_\confq = 
\begin{pmatrix} 
	g_{\theta^c}^T H^{-1} g_{\theta^c} 	&g_{\xi}^T H^{-1} g_{\theta^c} \\
	g_{\theta^c}^T H^{-1} g_{\xi}		&g_{\xi}^T H^{-1} g_{\xi} 
\end{pmatrix}, \nonumber
\end{align}
where we have denoted the partial Jacobians by
$g_{\theta^c} = \nabla^T_{\theta^c}g\in\R^{m\times n^c}$,
$g_{\xi} = \nabla^T_{\xi}g\in\R^{m\times n^\xi}$
and the matrix $H = g_{\theta^c}g_{\theta^c}^T + g_{\xi}g_{\xi}^T \in\R^{m\times m}$.

{\bf B component}.
Given $q_0,p_0\in T^*\manifold$ and a time $t>0$
\[
q_t = q_0,\qquad
p_t = p_0 - t \nabla_qV(q_0) - \nabla_q g(q_0)( \mu_t - \mu_0),
\]
where $\mu_t$ is such that $p_t\in T_{q_t}^*\manifold$ (i.e., it satisfies the constraint $0 = \nabla_q g(q_t)p_t$).
Note that as $q_0,p_0$ satisfy the constraints we have $\mu_0=0$.
Projecting onto the cotangent space $T_{q_t}^*\manifold = T_{q_0}^*\manifold$
and using $\Pi(q_0)\nabla_q g(q_0) = 0$ and $p_0=\Pi(q_0)p_0$, we obtain
\begin{align*}
p_t = \Pi(q_t)p_t = \Pi(q_0)\big( p_0 - t \nabla_qV(q_0) - \nabla_qg(q_0) \mu_t \big)\\
= p_0 - t \Pi(q_0)\nabla_qV(q_0).
\end{align*}

The B step is thus obtained for a chosen stepsize $h>0$ as:
given $q_n = (\theta^u_n,\theta^c_n,\xi_n)\in\manifold$
and \\ $p_n = (p^u_n,p^c_n,p^\xi_n)\in T_{q_n}^*\manifold$
\begin{equation}	\label{eq:UL_Bstep}
\text{(B, gen.)}\quad
\begin{aligned}
&\theta^u_{n+1} =\theta^u_{n},
\qquad
\theta^c_{n+1} = \theta^c_{n},
\qquad
\xi_{n+1} = \xi_{n},
\\[5pt]
&
p^u_{n+1} = p^u_{n} - h \nabla_{\theta^u} \loss(\theta_n),
\\[5pt]
&
\bar p^c_{n+1} = p^c_{n} - h \nabla_{\theta^c} \loss(\theta_n),
\qquad
\bar p^\xi_{n+1} = p^\xi_{n},
\\[5pt]
&
\begin{pmatrix} 
p^c_{n+1} \\  
p^\xi_{n+1}
\end{pmatrix}
= \Pi_{\confq}(\confq_n)
\begin{pmatrix} 
\bar p^c_{n+1}   \\
\bar p^\xi_{n+1}
\end{pmatrix}, \\
&\text{where } \confq_n = \begin{pmatrix} \theta^c_{n}\\ \xi_{n} \end{pmatrix}
\end{aligned}
\end{equation}

{\bf O component}.
Similarly as for the B part, the O part can be solved exactly in law for any constraint.
Given $q_0,p_0\in T^*\manifold$ and a time $t>0$, we have
\begin{align*}
q_t &= q_0,\\
p_t &= p_0 - \gamma \int_0^t p_t\,\d t + \sqrt{2\gamma \tau}\int_0^t\,\d \mathcal{W}_t -\nabla_qg(q_0) \nu_t,
\end{align*}
where $\nu_t$ ensures that $p_t\in T_{q_t}^*\manifold$.
Projecting to the cotangent space $T_{q_t}^*\manifold = T_{q_0}^*\manifold$ as before, we obtain
\begin{align*}
p_t &= \Pi(q_t)p_t \\
&= p_0 - \gamma \int_0^t  \Pi(q_0)p_t\,\d t + \sqrt{2\gamma \tau}\Pi(q_0)\int_0^t\,\d \mathcal{W}_t.
\end{align*}
We thus recognize that $p_t$ is an Ornstein--Uhlenbeck process:
\[
p_t \stackrel{\text{law}}{=}   \Pi(q_0) \big( e^{-\gamma t} p_0 + \sqrt{\tau(1-e^{-2\gamma t})} R \big)
\]
with $R \sim N(0,I_d)$, where the equality holds in law.

The O step is thus obtained for a chosen stepsize $h>0$ as:
given $q_n = (\theta^u_n,\theta^c_n,\xi_n)\in\manifold$
and $p_n = (p^u_n,p^c_n,p^\xi_n)\in T_{q_n}^*\manifold$
\begin{equation}	\label{eq:UL_Ostep}
\text{(O, gen.)}\quad
\begin{aligned}
&\theta^u_{n+1} =\theta^u_{n},
\qquad
\theta^c_{n+1} = \theta^c_{n},
\qquad
\xi_{n+1} = \xi_{n},
\\[5pt]
&
p^u_{n+1} =  e^{-\gamma h} p^u_{n} + \sqrt{\tau(1-e^{-2\gamma h})}R^u,
\\&
\bar p^c_{n+1} =  e^{-\gamma h} p^c_{n} + \sqrt{\tau(1-e^{-2\gamma h})}R^c,
\\&
\bar p^\xi_{n+1} = e^{-\gamma h} p^c_{n} + \sqrt{\tau(1-e^{-2\gamma h})}R^\xi,
\\[5pt]
&
\begin{pmatrix} 
p^c_{n+1} \\  
p^\xi_{n+1}
\end{pmatrix}
= \Pi_{\confq}(\confq_n)
\begin{pmatrix} 
\bar p^c_{n+1}   \\
\bar p^\xi_{n+1}
\end{pmatrix}
\\
&\text{where } \confq_n = \begin{pmatrix} \theta^c_{n}\\ \xi_{n} \end{pmatrix},
\end{aligned}
\end{equation}
and 
$R^u,R^c$, and $R^\xi$ are independent standard normal random variables.

\subsection{Circle constraint, overdamped Langevin (c-CoLod)}	\label{sec:discr_OLcircle}
We consider here the circle constraint Eq. (\ref{eq:circleConstraint}), for which the partial Jacobians are computed as
\begin{align}	\label{eq:jacobianG_circle}
\nabla_q^T g = \big(\nabla_{\theta^u}^Tg, \nabla_{\theta^c}^Tg, \nabla_{\xi}^T g\big) \in\R^{m\times(n^u+n^c+m)}, \nonumber  \\
\partial_{\theta^u_j} g_i = 0,
~~
\partial_{\theta^c_j} g_i = 2\theta^c_i\delta_{ij},
~~
\partial_{\xi_j} g_i = 2\xi_i\delta_{ij},
\end{align}
where $\delta_{ij}$ is the Kronecker delta. 

For this constraint, the projection step in Eq. \eqref{eq:OLdiscretization1} can be computed explicitly.
Indeed $\lambda_{n}$ can be found by solving the $m$ quadratic equations
$0=g_i(\bar{q}_{n+1} -\nabla_q g(q_n)\lambda_n)$ $1\leq i\leq m$.
The (potential) two roots of each equation corresponds to the (potential) two projections of $\bar{q}_{n+1}$ onto the circle following the direction $\nabla g_i(q_n) = 2(\theta^c_{n,i},\xi_{n,i})$. When two roots are found, we may select the one closest to the point of origin $(\theta^c_{n,i},\xi_{n,i})$. 
However, if the point to project $(\bar\theta^c_{n+1,i},\bar\xi_{n+1,i})$ is too far away from the circle, this oblique projection may not be possible (i.e., the quadratic equation has no real root).

For the circle constraint, method Eq. \eqref{eq:OLdiscretization2} thus leads to a more robust projection process.
Indeed, as $\nabla g_i(q_{n+1}) = 2(\theta^c_{n+1,i},\xi_{n+1,i})$, the direction of the projection is now orthogonal to the circle.
To find an expression for the orthogonal projection $P$ of a point $(\bar\theta_1, \bar\xi_1)$ on the circle, it is easier to use a geometrical approach than to find the Lagrange multipliers:
\[
(\theta_1,\xi_1) = P (\bar\theta_1, \bar\xi_1) = 
\big(r_i\cos(\alpha), r_i\sin(\alpha)\big),
\]
where $\alpha = \arctan\Big( \frac{\bar\xi_1}{\bar\theta_1}\Big)$.
We obtain the following discretization of the overdamped Langevin with circle constraints.
We initialize the parameters of the neural network using standard PyTorch initialization \cite{Pytorch, Pytorchinit}, i.e., $\mathcal{U}(-1/\sqrt{N_{in}},1/\sqrt{N_{in}})$, where $N_{in}$ is the number of inputs to a layer. The auxiliary variables $\xi_i$ corresponding to the constrained parameters $\theta^c_i$ are initialized to obey the constraint $(\theta^c_i)^2 + \xi^2_i = r^2_i$. 
For a chosen stepsize $h>0$ and given a configuration $q_n = (\theta^u_n,\theta^c_n,\xi_n)\in\manifold$,
one step of the method is defined by $q_{n+1} = (\theta^u_{n+1},\theta^c_{n+1},\xi_{n+1})\in\manifold$ as
\begin{equation}	\label{eq:OL_circle_discr}
\begin{aligned}
&\theta^u_{n+1,i} 
	= \theta^u_{n,i} - h\partial_{\theta^u_i} \loss(\theta_n) + \sqrt{2\beta^{-1} h} R^u_i, \\
&\bar\theta^c_{n+1,i} 
	= \theta^c_{n,i} - h\partial_{\theta^c_i}\loss(\theta_n) + \sqrt{2\beta^{-1} h} R^c_i, \\
&\bar\xi_{n+1,i} 		
	= \xi_{n,i} + \sqrt{2\beta^{-1} h} R^\xi_i, \\
&\alpha_{n,i}
	= \arctan \left( \frac{\bar\xi_{n+1,i}}{\bar\theta^c_{n+1,i}} \right), \\
&\theta^c_{n+1,i} 
	 = r_i\cos(\alpha_{n,i}),\\
&\xi_{n+1,i} 		
	= r_i\sin(\alpha_{n,i}),
\end{aligned}
\end{equation}
where 
$R^u_i,R^c_i, R^\xi_i$ are independent standard normal random variables.


\subsection{Circle constraint, underdamped Langevin (c-CoLud)}	\label{sec:discr_ULcircle}

We provide here the full discretization of the underdamped Langevin dynamics in the case of the circle constraint Eq. (\ref{eq:circleConstraint}).

\textbf{A component}.
For the circle constraint we can solve the A step explicitly.
First recall that as $\nabla^T_{\theta^u} g= 0$, the unconstrained parameters $\theta^u$ are obtained with a standard A step of the unconstrained underdamped Langevin.
Let us then focus on solving the constrained components: 
we denote $\confq=(\theta^c,\xi), p^\confq= (p^c,p^\xi)$.
Then for $1\leq i\leq m$ the A step in Eq. (\ref{eq:Bcomponent}) corresponds to the constrained ODEs
\begin{equation}	\label{eq:Acomponent_circle}
\begin{aligned}
&\dot{\confq}_i = p^\confq_i\\
&\dot{p}^\confq_i = -2 \lambda_i \confq_i\\
&|\theta^c_i|^2 + |\xi_i|^2 = r_i^2 ,\qquad
\theta^c_ip^c_i + \xi_i  p^\xi_i = 0.
\end{aligned}
\end{equation}
As these constrained ODEs are uncoupled, let us drop the specification of the index $i$.
By assumption, we are given initial conditions that satisfy the constraint $(\confq_0,p^\confq_0)\in T^*\manifold$.
Solving the second order ODE $\ddot{\confq} = -2\lambda \confq$, we find that any solution has the form
$\confq_t = R^{2\lambda}_t \confq_0$, where $R^{\omega}_t$ is a rotation matrix with angular speed $\omega$ given with its time derivative as
\begin{align*}
R^\omega_t &= \begin{pmatrix} \cos(\omega t) &\sin(\omega t) \\ -\sin(\omega t) & \cos(\omega t) \end{pmatrix},\\
\dot{R}^\omega_t &= \omega \begin{pmatrix} -\sin(\omega t) &\cos(\omega t) \\ -\cos(\omega t) & -\sin(\omega t) \end{pmatrix}.
\end{align*}
Computing the momentum $p^\confq_t = \dot{\confq}_t = \dot{R}^\omega_t \confq_0$,
and using the properties of $R^\omega_t$
we verify that $\confq_t,p^\confq_t$ satisfy the constraints in Eq. \eqref{eq:Acomponent_circle}
($\|.\|$ denotes the Euclidean norm in $\R^2$ and $\cdot$ the dot product):
\begin{align*}
&\|\confq_t\|^2 = \|R^\omega_t \confq_0\|^2 = \|\confq_0\|^2 = r^2, \\
&\confq_t\cdot p^\confq_t = \confq_0^T(R^\omega_t)^T \dot{R}^\omega_t\confq_0 = 0. 
\end{align*}
We still have to find the angular speed $\omega=2\lambda$ such that the momentum $p^\confq_t$ is consistent with its initial value $p^\confq_0$ (we denote $\confq_0=(\theta^c_0,\xi_0)$ and $p^\confq_0= (p^c_0,p^\xi_0))$:
\[
p^\confq_0 = \dot{R}^\omega_0 \confq_0 
\quad\Leftrightarrow\quad
p^c_0 = \omega \xi_0
~\text{ and }~
p^\xi_0 = -\omega \theta^c_0.
\]
We thus find that 
\begin{align*}
&\xi_0 p^c_0 -\theta^c_0 p^\xi_0 = \omega \big(|\xi_0|^2+|\theta^c_0|^2\big) = \omega r^2
\\
&\Leftrightarrow\quad
\omega = \frac{1}{r^2} \big(  \xi_0 p^c_0 - \theta^c_0 p^\xi_0 \big).
\end{align*}
We have thus found an explicit expression for the solution of the A component for circle constraints Eq. \eqref{eq:Acomponent_circle}.

To complete the B and O steps given in Eq. \eqref{eq:UL_Bstep} and Eq. \eqref{eq:UL_Ostep}, we need an explicit expression for the projection $\Pi_\confq$ in Eq. \eqref{eq:expressionPiw} (using Eq. \eqref{eq:jacobianG_circle}, recall that $m=n^c=n^\xi$):
\[
\Pi_\confq(w) = 
\begin{pmatrix}
I_m - D^{11} 			& - D^{12} \\
- D^{12}					&I_m - D_{22}
\end{pmatrix},
\]
where $D^{kl}\in\R^{m\times m}$ are the diagonal matrices defined as
\begin{align*}
D^{11}_{ii} &= \frac{|\theta^c_i|^2}{ |\theta^c_i|^2 + |\xi_i|^2 },
\quad
D^{12}_{ii} = \frac{\theta^c_i\xi_i}{ |\theta^c_i|^2 + |\xi_i|^2 },
\\
D^{22}_{ii} &= \frac{|\xi_i|^2}{ |\theta^c_i|^2 + |\xi_i|^2 }.
\end{align*}
Assuming that $\confq=(\theta^c,\xi)$ satisfies the constraint,
the projection of $(\bar{p}^c,\bar{p}^\xi)$ is thus computed as
\begin{align*}
&\begin{pmatrix}{p}^c \\{p}^\xi\end{pmatrix} 
	= \Pi_\confq (w)
		\begin{pmatrix}\bar{p}^c \\\bar{p}^\xi\end{pmatrix}
,\\
&\text{where} \ 
\begin{aligned}
{p}^c_i		
	&= \bar{p}^c_i   -\frac{\theta^c_i}{r_i^2}\big( \theta^c_i \bar{p}^c_i +\xi_i \bar{p}^\xi_i\big)
		\quad 1\leq i\leq m,\\
{p}^\xi_i 
	&= \bar{p}^\xi_i -\frac{\xi_i}{r_i^2}\big( \theta^c_i \bar{p}^c_i +\xi_i \bar{p}^\xi_i\big)
		\quad 1\leq i\leq m.
\end{aligned}
\end{align*}
Note that in the B step Eq. \eqref{eq:UL_Bstep}, the above expressions can be simplified 
by combining the simple definition of $(\bar{p}^c_n,\bar{p}^\xi_n)$ with the constraint 
\[
0 = \big(\nabla^Tg(q)p\big)_i = 2 \big(\theta^c_i p^c_i + \xi_i p^\xi_i \big).
\]

We provide below the explicit updates for the A, B and O components for circle constraints. We initialize the parameters of the net using standard PyTorch initialization \cite{Pytorch, Pytorchinit}. The auxiliary variables $\xi$ corresponding to the constrained parameters $\theta^c$ are initialized to obey the constraint $(\theta^c)^2 + \xi^2 = r^2$, so that $q_0 = (\theta^u_0,\theta^c_0,\xi_0)\in\manifold$. The momenta, $p^u, p^c$, and $p^{\xi}$, are generated in the same manner as for standard SGD with momentum in PyTorch, i.e., as equal to the initial gradients. Subsequently, the momenta belonging to the constrained variables $p^c$ and to the auxiliary variables $p^{\xi}$ are projected using $\Pi_\confq$, so that $p_0 = (p^u_0,p^c_0,p^\xi_0)\in T_{q_0}^*\manifold$.
For a stepsize $h>0$ we obtain
\begin{equation*}
\text{(A)} 
\left\{
\begin{aligned}
&\theta^u_{n+1,i} 
= \theta^u_{n,i} + h p^u_{n,i},
\quad
\\
&\omega_i = \frac{1}{r_i^2} \big(  \xi_{n,i} p^c_{n,i} - \theta^c_{n,i} p^\xi_{n,i} \big),
\\
&\theta^c_{n+1,i} 
= \cos(\omega_i h)\theta^c_{n,i}  + \sin(\omega_i h)\xi_{n,i}  , 
\\
&\xi_{n+1,i} 		
= -\sin(\omega_i h)\theta^c_{n,i}  + \cos(\omega_i h)\xi_{n,i} ,
\\[5pt]
&p^u_{n+1,i} = p^u_{n,i},
\\
&p^c_{n+1,i} = \omega_i\big(-\sin(\omega_i h)\theta^c_{n,i}  + \cos(\omega_i h)\xi_{n,i}\big),
\\
&p^\xi_{n+1,i} = -\omega_i\big(\cos(\omega_i h)\theta^c_{n,i}  +\sin(\omega_i h)\xi_{n,i}\big),
\end{aligned}
\right.
\end{equation*}
\begin{align*}
\text{(B)}
\left\{
\begin{aligned}
&\theta^u_{n+1} =\theta^u_{n},
\qquad
\theta^c_{n+1} = \theta^c_{n},
\qquad
\xi_{n+1} = \xi_{n},
\\[5pt]
&
p^u_{n+1} = p^u_{n} - h \nabla_{\theta^u} \loss(\theta_n),
\\&
\bar p^c_{n+1,i} = p^c_{n,i} - h\Big( 1 - \frac{1}{r_i^2}|\theta^c_{n,i}|^2 \Big) \partial_{\theta^c_i} \loss(\theta_n),
\\&
\bar p^\xi_{n+1,i} = p^\xi_{n,i} + h\frac{1}{r_i^2}\theta^c_{n,i}\xi_{n,i} \partial_{\theta^c_i} \loss(\theta_n),
\end{aligned}
\right.
\end{align*}

\begin{equation*}	
\text{(\text{O})} 
\left\{
\begin{aligned}
&\theta^u_{n+1} =\theta^u_{n},
\qquad
\theta^c_{n+1} = \theta^c_{n},
\qquad
\xi_{n+1} = \xi_{n},
\\[5pt]
&
p^u_{n+1} =  e^{-\gamma h} p^u_{n} + \sqrt{\beta^{-1}(1-e^{-2\gamma h})}R^u,
\\&
\bar p^c_{n+1} =  e^{-\gamma h} p^c_{n} + \sqrt{\beta^{-1}(1-e^{-2\gamma h})}R^c,
\\&
\bar p^\xi_{n+1} = e^{-\gamma h} p^c_{n} + \sqrt{\beta^{-1}(1-e^{-2\gamma h})}R^\xi,
\\
&
p^c_{n+1,i}	
=\Big(1 - \frac{|\theta^c_{n,i}|^2}{r_i^2}\Big)   \bar  p^c_{n+1,i}  
-  \frac{1}{r_i^2} \theta^c_{n,i}\xi_{n,i} \bar  p^\xi_{n+1,i},
\\
&
p^\xi_{n+1,i} 
=-\frac{\theta^c_{n,i}\xi_{n,i} \bar  p^c_{n+1,i}}{r_i^2} 
+ \Big(1 - \frac{|\xi_{n,i}|^2}{r_i^2}\Big) \bar  p^\xi_{n+1,i},
\end{aligned}
\right.
\end{equation*}
where $R^u,R^c$, and $R^\xi$ are vectors of independent standard normal random variables.

\subsection{Orthogonality constraint, overdamped Langevin dynamics (o-CoLod)}	\label{sec:OL_orthogonalConstraint}

We present here a particular discretization of the constrained overdamped Langevin dynamics Eq. (\ref{eq:constrainedOverdampedLangevin}) for the orthogonality constraint Eq. (\ref{eq:orthogonalConstraint}).

For notational convenience, we present the updates for the weight matrix $W^\ell$ of a given layer $\ell$.
The updates for the biases are standard Euler--Maruyama steps  such as given for $\theta^u$ in Eq. \eqref{eq:OL_circle_discr}.
 
Referring to Eq. (\ref{eq:orthogonalConstraint}), we denote
\begin{equation}
\begin{array}{llll}
Q = W^\ell,& r=n^{\ell}, &s = n^{\ell-1}	&\text{if }n^{\ell-1}\leq n^{\ell},\\
Q=(W^\ell)^T,& r=n^{\ell-1}, &s = n^{\ell}	&\text{otherwise}.
\end{array}
\end{equation}
so that $Q\in\R^{r\times s}$.
With this notation, the constraint Eq. (\ref{eq:orthogonalConstraint}) is $g(Q)=0$ where
\begin{equation}	\label{eq:orthogonalConstraint_g}
g : \R^{r\times s} \to \R^{s\times s}, \qquad g(Q) = Q^TQ - I_s.
\end{equation}
Recall that due to symmetry, the matrix equality $g(Q)=0_s$ corresponds to $s(s+1)/2$ constraints.
We compute the partial derivative
\begin{align}	\label{eq:partialDerivativeOGconstraint}
&\partial_{Q_{kl}} g_{ij}(Q) = \delta_{li} Q_{kj} + \delta_{lj} Q_{ki} \\
&1\leq i,j,k\leq s, ~1\leq l\leq r. \nonumber
\end{align}
In particular, if $\Lambda$ is an $s\times s$ symmetric matrix, we verify that 
\[
\sum_{i,j=1}^s \partial_{Q_{kl}} g_{ij}(Q) \Lambda_{ij}  = 2\big( Q\Lambda\big)_{kl}.
\]
We thus obtain the natural matrix form of the constrained dynamics Eq. (\ref{eq:constrainedOverdampedLangevin}):
$Q_t:(0,\infty)\to \R^{r\times s}$ solves
\begin{equation}	\label{eq:continuousSDE_orthogonalConstraint}	
\begin{aligned}
&\d Q_t = -\nabla_Q V(Q_t) \,\d t + \sqrt{2\beta^{-1}}\,\d \mathcal{W}_t  - Q_t \,\d \Lambda_t ,\\
&g(Q_t) = 0,
\end{aligned}
\end{equation}
where
$\big(\nabla_Q V\big)_{ij} = \partial_{Q_{ij}} V = \partial_{W^\ell_{ij}} \loss$ (or $\partial_{W^\ell_{ji}} \loss$)
and $\mathcal{W}_t$ is a Wiener process in $\R^{r\times s}$.
Furthermore the process $\Lambda_t$ has values in the $s\times s$ symmetric matrices and is the Lagrange multiplier corresponding to the $s(s+1)/2$ constraints.

Applying discretization scheme Eq. \eqref{eq:OLdiscretization1} to Eq. \eqref{eq:continuousSDE_orthogonalConstraint}, we obtain the iteration step
$Q_n\in\manifold \mapsto Q_{n+1}\in\manifold$ given by
\begin{equation}	\label{eq:OLdiscr_orthogonalConstraint}
\begin{aligned}
&\bar Q_{n+1} = Q_n - h\nabla_Q V(Q) + \sqrt{2\beta^{-1}h} R_n,\\
&Q_{n+1} = \bar{Q}_{n+1} - Q_n\Lambda_n,
\end{aligned}
\end{equation}
where $\Lambda_n$ is a symmetric $s\times s$ matrix s.t. $g(Q_{n+1})=0$
and $R_n\in\R^{r\times s}$ is a matrix of independent standard normal random variables.

Note that the projection step in Eq. \eqref{eq:OLdiscr_orthogonalConstraint} requires to solve a non-linear system.
Following a similar technique as described in \cite{LeR04}[Chap. 8], we derive a quasi-Newton scheme for that task. 
Using the fact that $Q_n$ satisfies the constraint we verify that
\[
\bar{Q}_ {n+1}^T Q_n = I_s - h\nabla_Q V(Q_n)^TQ_n + \sqrt{2\beta^{-1}h} R_n^TQ_n.
\]
The constraint $g(Q_{n+1})=0$ thus reads
\begin{align}	\label{eq:estimateQuasiNewton}
0 	
	= \big(\bar{Q}_{n+1} - Q_n\Lambda_n\big)^T \big(\bar{Q}_{n+1} - Q_n\Lambda_n\big)-I_s \nonumber \\
	= \big(\bar{Q}_{n+1}^T\bar{Q}_{n+1} - I_s\big) - 2\Lambda_n + \mathcal{O}(\sqrt{h}),
\end{align}
where $\mathcal{O}(\sqrt{h})$ denotes a matrix whose 2-norm has order $\sqrt{h}$.
Solving for $\Lambda_n$, we find
\[
\Lambda_n = \frac{1}{2} \big( \bar{Q}_{n+1}^T\bar{Q}_{n+1} - I_s \big)  + \mathcal{O}(\sqrt{h}).
\]
Neglecting the terms of order $\sqrt{h}$ and higher, we obtain the following quasi-Newton scheme:
setting $Q^{(0)}=\bar{Q}_{n+1}$, repeat the iteration
\begin{align}	\label{eq:quasiNewton_orthogonalConstraint}
&Q^{(k+1)} = Q^{(k)} - Q_n\Lambda^{(k)},  \\
&\text{where } \Lambda^{(k)} = \frac{1}{2} \big( (Q^{(k)})^TQ^{(k)} - I_s \big), \nonumber
\end{align}
until the process reaches convergence and set $Q_{n+1} = Q^{(k+1)}$.
To assess whether convergence has been reached, a tolerance on the $2$-norm of $\Lambda^{(k)}$ can be assigned:
$\|\Lambda^{(k)}\| \leq \mathrm{TOL}$.
However in practice, to ensure that the process ends and to avoid undesirable overhead we typically prefer to either combine this stopping criterion with a limit for the number $K$ of iterations, or use a fixed number of iterations $K$.
Note that estimate Eq. \eqref{eq:estimateQuasiNewton} ensures that a small number of iterations $K$ is sufficient for the constraint to be satisfied up to a small error.

The initialization for the constrained weights is performed following \cite{Saxe2013}, which is an built-in option in PyTorch. Other parameters are initialized using the standard PyTorch initialization \cite{Pytorch, Pytorchinit} unless otherwise indicated. Constraints are applied layer-wise, where for convolutional layers with weight tensors of the size $n_l \times n_{l-1} \times n_h \times n_w$ (where $n_h$ and $n_w$ are the height and width of the kernel) the weight matrices are reshaped as $n_l \times n_{l-1} n_h n_w$. For CNNs these reshaped matrices are typically rectangular. If they are thin, but long (i.e., $n_l > n_{l-1} n_h n_w$) we apply the constraint $W^T W = I$, but if they have more columns than rows we apply the constraint $W W^T = I$. 

\subsection{Orthogonality constraint, underdamped Langevin (o-CoLud)}	\label{sec:UL_orthogonalConstraint}

To discretize the underdamped Langevin constrained dynamics, we need the orthogonal projection $\Pi$ onto the cotangent space $T_Q^*\manifold$. As the constraint Eq. \eqref{eq:orthogonalConstraint_g} is given in a matrix form, using the formula Eq. \eqref{eq:OL_definitionPi} is not very convenient so we will rather derive $\Pi$ from its projection property.

Using Eq. \eqref{eq:partialDerivativeOGconstraint}, we find that for $1\leq i,j\leq s$
\[
0 = \sum_{k=1}^s\sum_{l=1}^r \partial_{Q_{kl}} g_{ij}(Q) P_{kl} = (P^TQ + Q^TP)_{ij},
\]
which leads to the following convenient expression for the cotangent space 
\[
T^*_Q\manifold = \big\{  P \in\R^{r\times s} \mid P^TQ + Q^TP = 0_s \big\}.
\]
Now, given $\bar P\in\R^{r\times s}$ we want to find a symmetric $s \times s$ matrix $\Lambda$ such that 
$P = \bar P - Q\Lambda$ belongs to $T^*_Q\manifold$, i.e.,
\[
0_s = P^TQ -Q^TP =  \bar P^TQ + Q^T\bar P - \Lambda Q^TQ - Q^TQ \Lambda.
\]
This equation is easily solved for $Q\in\manifold$ and we find 
$\Lambda = \frac{1}{2} (\bar P^TQ + Q^T\bar P)$. We obtain the following expression for the projection onto the cotangent space:
\begin{align*}
&\Pi_Q : \R^{r\times s} \to \R^{r\times s} ,\\
&\bar P \mapsto \Pi_Q\bar P = \bar P - \frac{1}{2} Q (\bar P^TQ + Q^T\bar P).
\end{align*}
We then verify that $\Pi_Q$ is indeed a projection onto the cotangent space $T_Q^*\manifold$
(i.e., $\Pi_Q\bar P\in T_Q^*\manifold$ $\forall\bar P\in\R^{r\times s}$ and $\Pi_Q^2 = \Pi_Q$) 
and that this projection is orthogonal with respect to the Frobenius inner product on $\R^{r\times s}$ 
(i.e., $\langle \bar P-\Pi_Q\bar P, P \rangle = 0$, where $\langle A,B\rangle = \mathrm{tr}(A^TB)$).
\ \\ \ \\ 
\textbf{A component}.
For the orthogonal constraint, the A component in Eq. (\ref{eq:Bcomponent}) can only be solved approximately.
A simple yet efficient discretization of A is the RATTLE scheme (see e.g. \cite{LeR04}[Chap. 8]):
\begin{equation}	\label{eq:ogConstraint_A_RATTLE0}
\begin{aligned}
&Q_{n+1} = Q_n  + hP_{n+1/2},\\
&P_{n+1/2} = P_n - Q_n\Lambda_{n+1/2}
\\	&\quad \text{where $\Lambda_{n+1/2}$ is s.t. }Q_{n+1}^TQ_{n+1} = I_s,\\
&P_{n+1} = P_{n+1/2} - Q_{n+1} \Lambda_{n+1} \\
	&\quad \text{where $\Lambda_{n+1}$ is s.t. }Q_{n+1}^TP_{n+1} + P_{n+1}^T Q_{n+1} = 0_s.
\end{aligned}
\end{equation}
Denoting $\bar\Lambda_{n+1/2} = h\Lambda_{n+1/2}$, $\bar P_{n+1}=P_{n+1/2}$ and using the projection operator $\Pi_Q$, Eq. \eqref{eq:ogConstraint_A_RATTLE0} can be rewritten as  
\begin{equation}
\begin{aligned}
&\bar Q_{n+1} = Q_n  + hP_{n},\\
&Q_{n+1} = \bar Q_{n+1} - Q_n\bar\Lambda_{n+1/2}\\
&\quad \text{where $\bar\Lambda_{n+1/2}$ is s.t. }Q_{n+1}^TQ_{n+1} = I_s 
\quad \text{(use Eq. \eqref{eq:quasiNewton_orthogonalConstraint})},\\
&\bar P_{n+1} =P_n - \frac{1}{h} Q_n\bar\Lambda_{n+1/2},\qquad
P_{n+1} = \Pi_{Q_{n+1}} \bar P_{n+1}.
\end{aligned}
\end{equation}
As in the overdamped case, we may now use the quasi-Newton scheme Eq. \eqref{eq:quasiNewton_orthogonalConstraint} for the projection step (to approximate $\bar\Lambda_{n+1/2}$).
Using $K$ iterations of the quasi-Newton scheme Eq. \eqref{eq:quasiNewton_orthogonalConstraint} (i.e., $Q_{n+1}=Q^{(K)}$),
we verify that $-Q_n\bar\Lambda_{n+1/2}$ satisfies
\begin{align*}
- Q_n\bar\Lambda_{n+1/2} 
	&= \sum_{k=0}^{K-1} Q_n\Lambda^{(k)} = \sum_{k=0}^{K-1} Q^{(k+1)}- Q^{(k)}\\
	&= Q^{(K)}-Q^{(0)}
	= Q_{n+1} - \bar Q_{n+1},
\end{align*}
so that $\bar P_{n+1} =P_n + \frac{1}{h}(Q_{n+1} - \bar Q_{n+1})$.

We obtain the following full discretization of the underdamped Langevin dynamics with orthogonality constraint.
The initialization for the constrained weights is performed following \cite{Saxe2013}. Corresponding momenta are initialized as the initial gradients (equivalently to standard PyTorch initialization) and subsequently projected using $P_0 = \bar{P}_0 - \frac{1}{2}Q_0(\bar{P}^T_0 Q_0 + Q_0^T \bar{P}_0)$. The A,B,O steps are then given as:

\begin{equation*}	
\text{(\text{A, OG})}
\left\{
\begin{aligned}
&\bar Q_{n+1} = Q_n  + hP_{n},\quad Q^{(0)} = \bar Q_{n+1}, \\
&\text{for $k = 0$ to $K-1$:} \\
&\qquad Q^{(k+1)} = Q^{(k)} - Q_n\Lambda^{(k)}, \\
&\quad\text{where } \Lambda^{(k)} = \frac{1}{2} \Big( \big(Q^{(k)}\big)^T Q^{(k)} - I_s \Big), \\
&Q_{n+1} = Q^{(K)},\\
&\bar P_{n+1} =P_n + \frac{1}{h}\big(Q_{n+1}-\bar Q_{n+1}\big), \\
&P_{n+1} = \Pi_{Q_{n+1}} \bar P_{n+1} = \bar{P}_{n+1} \\
&\ \ \ - \frac{1}{2}Q_{n+1} \Big(\bar{P}^T_{n+1} Q_{n+1} + \big(Q_{n+1}\big)^T \bar{P}_{n+1}\Big).
\end{aligned}
\right.
\end{equation*}
\begin{equation*}
\text{(B, OG)}
\left\{
\begin{aligned}
&Q_{n+1} =Q_{n},
\\
&\bar P_{n+1} = P_n - h\nabla_Q V(Q_n),\quad
\\&P_{n+1} = \Pi_{Q_n}P_{n+1} \\
	&= \bar{P}_{n+1}  - \frac{1}{2}Q_{n} \Big(\bar{P}^T_{n+1} Q_{n} + \big(Q_{n}\big)^T \bar{P}_{n+1}\Big),
\end{aligned}
\right.
\end{equation*}
\begin{equation*}	
\text{(\text{O, OG})}
\left\{
\begin{aligned}
&Q_{n+1} =Q_{n},
\\
&\bar P_{n+1} = e^{-\gamma h} P_{n} + \sqrt{\beta^{-1}(1-e^{-2\gamma h})}R_n,
\\&P_{n+1} = \Pi_{Q_n} \bar P_{n+1} \\
 	&=  \bar{P}_{n+1}  - \frac{1}{2}Q_{n} \Big(\bar{P}^T_{n+1} Q_{n} + \big(Q_{n}\big)^T \bar{P}_{n+1}\Big),
\end{aligned}
\right.
\end{equation*}
where $R_n$ is a matrix of independent standard normal random variables.

\newpage
\section{Feedforward neural network notations and gradients (backpropagation)}	\label{sec:nnGradient}

Given a dataset $X= \{x_i,y_i\}_{i=1}^N$, where $x_i\in\R^{d^\mathrm{in}}, y_i\in\R^{d^{\mathrm{out}}}$, we want to construct an interpolant of the relation $x_i \mapsto y_i$.
For this task, we choose a feedforward neural network (NN) with $L+1$ layers (i.e., $L$ parametrized layers, $L$ is the \textit{depth}).
For $1\leq \ell\leq L$ we denote the \textit{width} of layer $\ell$ as $d^{\ell}$ ($d^0 = d^\mathrm{in}$, $d^L=d^{\mathrm{out}}$).	
The parameters of the NN at layer $\ell$ are given by the weights and biases
\[
W^{\ell} \in\R^{d^\ell\times d^{\ell-1}},
\quad
b^\ell \in\R^{d^{\ell}}
\qquad
1\leq \ell \leq L.
\]
For notational convenience, let us stack the parameters in a vector
\begin{align*}
\theta^\ell &= \begin{pmatrix} \theta^\ell_W\\\theta^\ell_b\end{pmatrix},
\quad \theta^\ell_b = b^\ell\in\R^{d^{\ell}}, \\
\theta^\ell_W &= \mathrm{vect}(W^\ell) 
= \begin{pmatrix} W^\ell e_1\\ \vdots \\ W^\ell e_{d^{\ell-1}} \end{pmatrix}\in\R^{d^{\ell}d^{\ell-1}}.
\end{align*}
In particular $\theta^\ell\in\R^{n^\ell}$, where $n^\ell$ is the number of parameters in layer $\ell$, 
$n^\ell = d^\ell \times d^{\ell-1} + d^\ell$.
The vector of all parameters is denoted
$\theta = (\theta^1,\hdots,\theta^\ell)\in\R^{n}$, where $n = \sum_{l=1}^L n_\ell$.

Each layer $1\leq \ell\leq L$ is equipped with an \textit{activation function} $\varphi^{\ell} : \R^{d^\ell}\to\R^{d^\ell}$,
which is is applied component wise: $\varphi^\ell_i(x) = \phi^\ell(x_i)$, for some $\phi^\ell:\R\to\R$.
In each layer $1\leq \ell \leq L$, we define the following functions
\begin{alignat*}{2}
&a^{\ell}: \R^{n^\ell\times d^{\ell-1}} \to \R^{d^{\ell}},
\quad
&&a^{\ell}(\theta^\ell,z^{\ell-1}) = W^\ell z^{\ell-1} + b^{\ell},
\\
&z^{\ell}: \R^{n^\ell\times d^{\ell-1}}\to \R^{d^{\ell}},
\quad
&&z^{\ell}(\theta^\ell,z^{\ell-1}) =\varphi^\ell(a^{\ell}(\theta^\ell,z^{\ell-1})),
\end{alignat*}
to which we associate the following shorthand notation
\begin{align*}
a^{\ell}_{\theta^\ell} &= a^{\ell}(\theta^\ell,\cdot) : \R^{d^{\ell-1}} \to \R^{d^{\ell}},
\\
z^{\ell}_{\theta^\ell} &= z^{\ell}(\theta^\ell,\cdot) : \R^{d^{\ell-1}} \to \R^{d^{\ell}}.
\end{align*}
We verify that the map $\theta^\ell\mapsto a^{\ell}(\theta^\ell,z^{\ell-1})$ can be written as
\begin{align*}
a^{\ell}(\theta^\ell,z^{\ell-1})
&= \big( (z^{\ell-1})^T \otimes I_{d^\ell}\big)\theta^\ell_W + \theta^\ell_b  \\
&= \big( (z^{\ell-1})^T \otimes I_{d^\ell}, I_{d^\ell}\big) \theta^\ell,
\end{align*}
where $I_{d}$ denotes the identity matrix in $\R^{d}$ and for $z\in\R^{s}$, $z^T \otimes I_d = (z_1I_d , \hdots,z_{s}I_d \big)$.
We then introduce the \textit{intermediate classifiers} as
$p^0(x) = x$ and
\[
p^\ell: \R^{n \times d^{0}} \to \R^{d^{\ell}},
\quad 
p^\ell(\theta,x) =
z^{\ell}_{\theta^\ell}
\circ\ \cdots \ \circ 
z^1_{\theta^1}(x),
\]
$1\leq \ell\leq L$, for which we use the shorthand $p^\ell_\theta = p^\ell(\theta,\cdot)$.
The (final) classifier is then the function $p_\theta = p^L_\theta: \R^{d^\mathrm{in}} \to \R^{d^\mathrm{out}}$.

To train the NN on the dataset $X$, we define the \textit{loss function} as
\[
\loss:\R^{n}\to\R\qquad
\loss(\theta) = -\sum_{i=1}^N D\big( p(\theta,x_i) , y_i \big),
\]
where $D = D(\hat y,y): \R^{d^\mathrm{out}}\times \R^{d^\mathrm{out}}\to\R$ is a function that measures the discrepancy between $\hat y$ and $y$.
In a simple classification case, $d^\mathrm{out}=1$ and $D$ is chosen to be the cross-entropy.
All the commonly used training method require the computation of the gradient of the loss function given as
\begin{align*}
&\nabla_\theta \loss:\R^{n}\to\R^{n}\\
&\nabla_\theta \loss(\theta) 
= - \sum_{i=1}^N \nabla_{\hat y} D\big( p(\theta,x_i) , y_i \big) \nabla_\theta p(\theta,x_i).
\end{align*}

\subsection*{Expression for the gradient of the loss (backpropagation)}

Recall that we denote the Jacobian matrix of a function $f:\R^n\to\R^m$ as the map $\nabla^T f:\R^n\to\R^{m\times n}$ defined as $(\nabla^T f)_{ij} = \partial_j f_i$.
Given two functions 
$f:\R^{m^1}\to\R^{m^2}$ and 
$g:\R^{m^2}\to\R^{m^3}$ the chain rule implies the Jacobian matrix of the composition $g\circ f$  satisfies
\begin{align*}
&\nabla^T (g\ \circ f)
: \R^{m^1} \to \R^{m^3\times m^1},
\\
&x\mapsto \nabla^T (g\ \circ f)(x) = \nabla^T g\big( f(x)\big)\nabla^T f(x).
\end{align*}

We compute the partial Jacobians of $a^j(\theta^j,z^{j-1})$ as
\begin{align}	\label{eq:grad_theta_aj}
&\nabla_{\theta^j}^T a^j:\R^{n^j}\times \R^{d^{j-1}} \to \R^{d^j\times n^j}, \nonumber \\
&\nabla_{\theta^j}^T a^j(\theta^j,z^{j-1}) 
=	\big( (z^{j-1})^T \otimes I_{d^j} , I_{d^j} \big),
\end{align}
and 
\begin{align}	\label{eq:grad_z_aj}
&\nabla_{z^{j-1}}^T a^j:\R^{n^j}\times \R^{d^{j-1}} \to \R^{d^j\times d^{j-1}} \nonumber \\
&\nabla_{z^{j-1}}^T a^j(\theta^j,z^{j-1}) = W^j .
\end{align}
The partial Jacobians of $z^j(\theta^j,z^{j-1})$ are then
\begin{equation}	\label{eq:partJacobZj}
\begin{aligned}
&\nabla_{\theta^j}^T z^j
	:\R^{n^j}\times \R^{d^{j-1}} \to \R^{d^j\times n^j}, \\
&\nabla_{\theta^j}^T z^j(\theta^j,z^{j-1}) 
	= \nabla_{a^j}^T\varphi^j\big( a^j_{\theta^j}(z^{j-1}) \big) \nabla_{\theta^j}^T a^j(\theta^j,z^{j-1}),\\
&\nabla_{z^{j-1}}^T z^j
	:\R^{n^j}\times \R^{d^{j-1}} \to \R^{d^j\times d^{j-1}}, \nabla_{z^{j-1}}^T  z^j(\theta^j,z^{j-1}) \\
	&= \nabla_{a^j}^T\varphi^j\big( a^j_{\theta^j}(z^{j-1}) \big) \nabla_{z^{j-1}}^T a^j(\theta^j,z^{j-1}),
\end{aligned}
\end{equation}
where we note that $\big(\nabla_{a^j}\varphi^j(z)\big)_{rs} = \partial_t \phi^j (z_r)\delta_{rs}$ (i.e., the matrix is diagonal).

The partial Jacobians of the classifier are then given by
\begin{equation}	\label{eq:backpropagation}
\begin{aligned}
\nabla_{\theta^{\ell}}^T p(\theta,x)
&= 	\nabla_{z^{L-1}}^T z \big(\theta^L, p^{L-1}_\theta(x) \big)
\cdots \\
&\cdots
\nabla_{z^{\ell}}^T z^{\ell+1} \big(\theta^{\ell+1}, p^{\ell}_\theta(x) \big)
\nabla_{\theta^{\ell}}^T z^{\ell} \big(\theta^{\ell}, p^{\ell-1}_\theta(x) \big), \\
&\quad 1\leq \ell\leq L-1,
\\
\nabla_{\theta^L}^T p(\theta,x) &= \nabla_{\theta^{L}}^T z^{L} \big(\theta^{L}, p^{L-1}_\theta(x) \big) ,
\end{aligned}
\end{equation}
and
\begin{align}	\label{eq:gradientxClassifier}
\nabla_{x}^T p(\theta,x)
&= 	\nabla_{z^{L-1}}^T z^L \big(\theta^L, p^{L-1}_\theta(x) \big) \nonumber \\
&\cdots 
\nabla_{z^{1}}^T z^{2} \big(\theta^{2}, p^{1}_\theta(x) \big)
\nabla_{x}^T z^{1} \big(\theta^{1}, x \big) .
\end{align}
From \eqref{eq:gradientxClassifier}, replacing the partial Jacobians of $z^j$ with the expressions provided in \eqref{eq:partJacobZj}, we obtain
\begin{equation}	
\nabla_{x}^T p(\theta,x)
= 	F_x^L W^L\cdots F_x^{2} W^{2} F_x^{1} W^1,
\end{equation}
where $F_x^j$ is the Jacobian matrix of the activation in the $j$th layer, $\varphi^{j}$ (e.g., if $\varphi^{j}=\mathrm{ReLU}$, $F_x^j$ is a diagonal matrix with 1 and 0 entries).
Constraining the weights moreover has a direct influence on the smoothness of the interpolant $p_\theta(x)$. 

From \eqref{eq:backpropagation}, replacing the partial Jacobians of $z^j$ with the expressions provided in \eqref{eq:partJacobZj},
we obtain 
\begin{align}	\label{eq:partialJacobian}
\nabla_{\theta^L}^T p_\theta(x)
&= 	F_x^L P_x^{L}, \nonumber \\
\nabla_{\theta^\ell}^T p_\theta(x)
&= 	F_x^L W^L\cdots F_x^{\ell+1} W^{\ell+1} F_x^{\ell} P_x^{\ell}, \nonumber \\ 
&\ \ \ \ \ \ 1\leq \ell \leq L-1,
\end{align}
where matrices $F_x^j$ are defined above
and $P_x^{j}$ is sparse with repeated entries of 
$p_\theta^j(x) = z^{j}_{\theta^j} \circ\cdots\circ z^1_{\theta^1}(x)$. 
This shows that as the depth $L$ is increased, the gradient of $p_\theta(x)$ with respect to the parameters of any
layer is composed of sparse products of the weights $W^j$.
This multiplicative structure leads to difficulty of DNN training:
the multiplication of small weights $\ll1$ leads to a low value of the gradient which in turn has the effect of slowing the training (\textit{vanishing gradient}), while
the multiplication of large weights $\gg1$ leads to a large value of the gradient which affects the stability of the learning procedure (\textit{exploding gradient}).
%

Let us explain the stability in more detail.
As training methods are discretization of a dynamics involving the gradient $\nabla_\theta \loss$, the stability of a method is connected to the Lipschitz constant $L$ on the statespace $E=\R^{n}$ of the gradient.\footnote{Recall that the Lipschitz constant of a function $h:E\subset R^r\to\R^s$ is the smallest constant $M$ such that
	$|f(x)-f(y)| \leq M|x-y|$ for all $x,y\in E$, where $|.|$ denotes the Euclidean norm.}
Assuming that $\loss$ is twice differentiable, the largest $L$ can be is
\begin{equation}	\label{eq:lipschitz}
M \leq \sup_{\theta\in E} \big|\lambda_{\max}(\theta)\big|,
\end{equation}
where $\lambda_{\max}(\theta)$ denotes the largest eigenvalue of the Hessian $\nabla_\theta^2 \loss(\theta)$.
The entries of the Hessian are computed as
\[
\begin{aligned}
\big(\nabla_\theta^2 & \loss(\theta)\big)_{rs}
= \\
&\sum_{i=1}^N \Big(\nabla_\theta p(\theta,x_i) \nabla_{\hat y}^2 D\big( p(\theta,x_i) , y_i \big) \nabla_\theta^T p(\theta,x_i)\Big)_{rs}
\\
&\ \ +\sum_{k=1}^{d^\mathrm{out}}\partial_{\hat y_k} D\big( p(\theta,x_i) , y_i \big) \partial^2_{\theta_r \theta_s} p_k(\theta,x_i).
\end{aligned}
\]
Even without providing the heavy expression of $\partial^2_{\theta_r \theta_s} p_k$,
using \eqref{eq:backpropagation} in this expression allow to appreciate the
impact of the magnitudes of the weights and of the depth on the Hessian and thus on the stability.

\section{Additional Numerical Details and 
Results}\label{sec:NumericsAppx}
We perform all experiments using PyTorch \cite{Pytorch} on NVIDIA DGX-1 GPUs. We compare our constrained methods with PyTorch's SGD with momentum optimiser. Unless otherwise indicated, we use for SGD $h = 0.1$ and $mom = 0$ (to compare with our constrained overdamped Langevin method) or $mom = 0.9$ (to compare with our constrained underdamped Langevin method). We use standard PyTorch initialization for all unconstrained parameters \cite{Pytorchinit, Pytorch}. Below we provide implementation details for all our experiments. 

\subsection{Orthogonality Constraints}
\begin{figure}[h]
\vskip -0.1em
    \centering
    \includegraphics[scale=0.4]{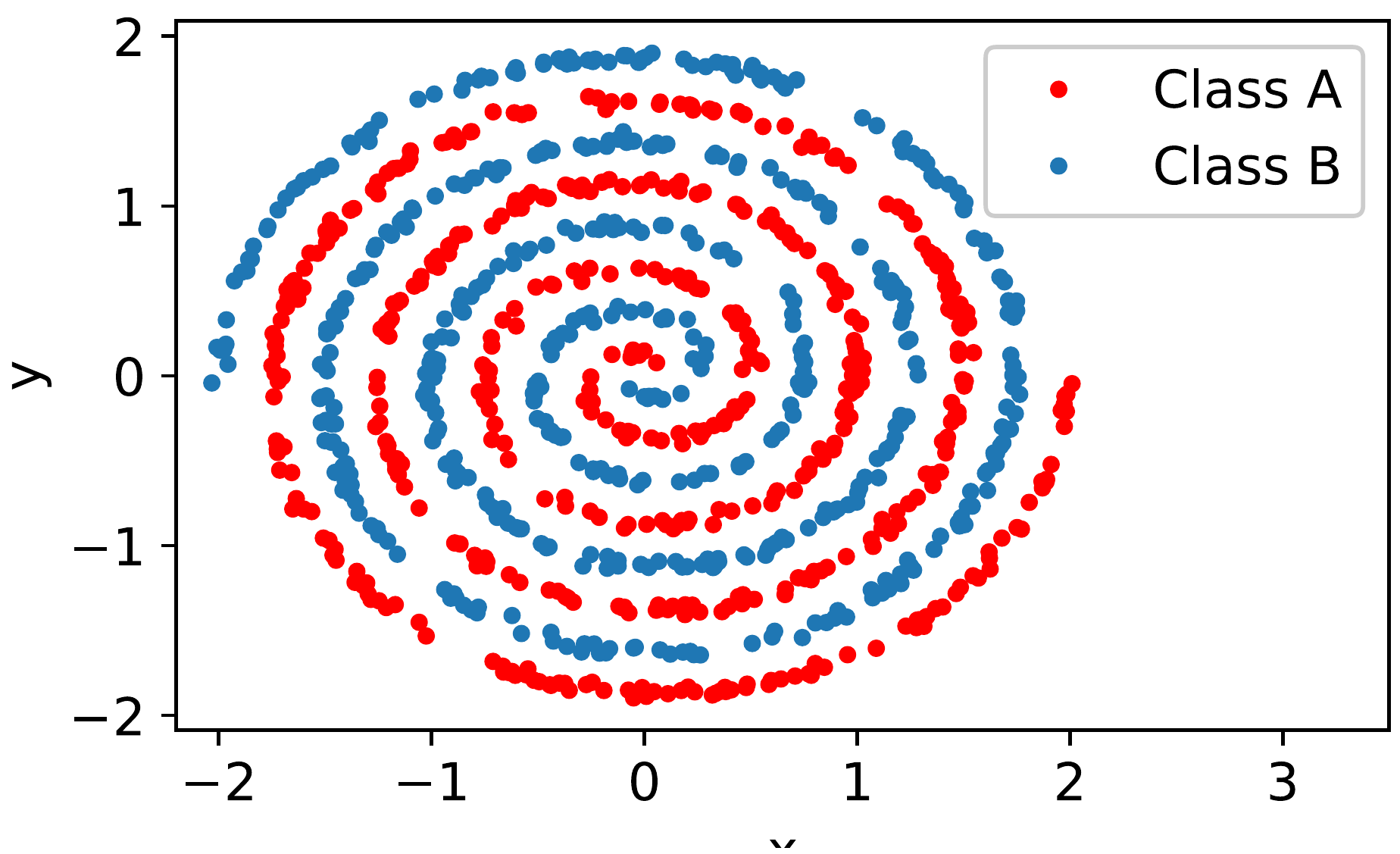}\caption{Data set generated using Eq. \eqref{spiraleqn}.} \label{Spiraldata}
\end{figure}

A plot of the planar spiral data set binary classification problem as used to produce Figure \ref{MLP_OCSGD} and Figure \ref{Stiffness} is provided in Figure D\ref{Spiraldata}. The first class of the data set is generated using 
\begin{align}
 x& = 2 \sqrt{t} \cos(8 \sqrt{t} \pi) +0.02\mathcal{N}(0,1), \nonumber \\
 y& = 2 \sqrt{t} \sin(8 \sqrt{t} \pi) +0.02\mathcal{N}(0,1), \label{spiraleqn}
\end{align}
where $t$ is drawn repeatedly from the uniform distribution $\mathcal{U}(0,1)$ to generate data points. The other class of this dataset is obtained by shifting the argument of the trigonometric functions by $\pi$. For our experiments we used 500 training data, 1000 test data points and 5\% subsampling. 

\makeatletter
\renewcommand{\fnum@figure}{\figurename~D\thefigure}
\makeatother

\begin{figure*}
    \centering
\makeatletter
\renewcommand{\fnum@figure}{\figurename~D\thefigure}
\makeatother
    \includegraphics[scale=0.45]{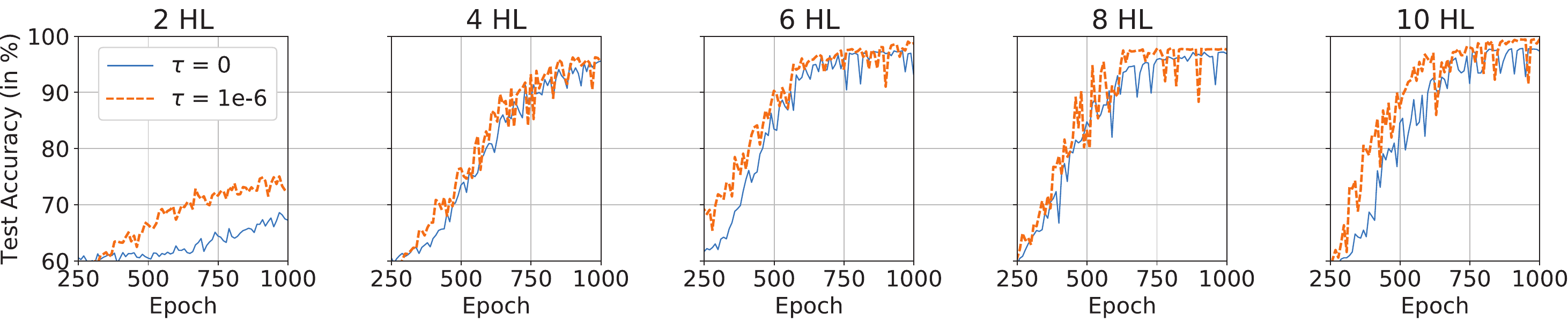}
    \caption{The effect of temperature on the performance of the o-CoLA-od optimizer for the 4-turn spiral data set (same set-up as for Fig. \ref{MLP_OCSGD}). MLPs with varying numbers of hidden layers (HL) were trained using o-CoLA-od with $h = 0.1$ and either $\tau = 0$ (blue line) or $\tau$ = 1e-6 (orange line). Results are averaged over 5 runs. The use of temperature is shown to speed up training and often slightly increases the obtained test accuracies.}\label{Spiral_withT} 

     \centering
       \includegraphics[clip, trim=0cm 6cm 0cm 4.5cm,width=0.6\linewidth]{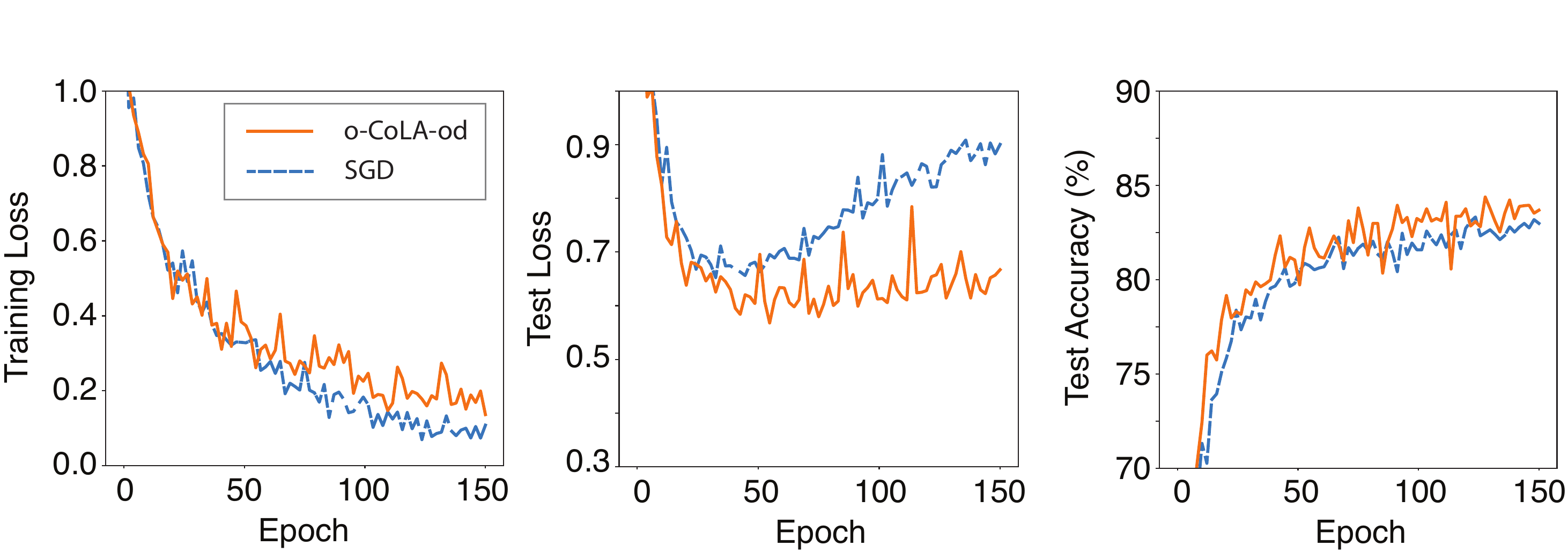}
       \vspace{-1mm}
     \caption{Training loss (left), test loss (middle) and test accuracy (right) of a ResNet-34 trained using SGD vs. o-CoLA-od on CIFAR-10 data, $h = 0.1$ (averaged over 5 runs). The orthogonality constraint provides modestly higher test accuracy and inhibits overfitting.\label{CSGD_Resnet}}
\end{figure*}

To generate the results presented in Figure \ref{MLP_OCSGD} and \ref{Stiffness} of the main paper, which show the effect of orthogonality constraints on this spiral data set (see Fig. D\ref{Spiraldata}), we use multi-layer perceptrons with ReLU activation and binary cross entropy (BCE) loss. In our experiments we vary the number of 100-node hidden layers of the multi-layer perceptrons. To compare the performance of our o-CoLod constrained method with standard SGD we set the temperature $\tau = 0$ and $h = 0.1$ for all methods to generate Fig. \ref{MLP_OCSGD}. For Fig. \ref{Stiffness} we do a grid-search to find the optimal value of the penalty strength for the orthogonal regularization approach with respect to the stepsize.
In Fig. D\ref{Spiral_withT} we show the effect of using a small temperature perturbation $\tau$ = 1e-6. 
The size of the temperature parameter was chosen to approximately match observed fluctuations in the loss function. A more precise parameterization is left for a subsequent work.


We also applied our orthogonality-constrained methods to the ResNet-34 architecture on CIFAR-10 image classification data \cite{cifar10}, see Figure \ref{OG-OBA_Resnet}. The input data is pre-processed using random crop (pad=4), random horizontal flip, and normalization. In this setting, running SGD with orthogonal initialization worsened the generalization performance of the resulting net and hence the standard PyTorch initialization was used for SGD. We train for 150 epochs and use a batchsize of 128. In Figure D\ref{CSGD_Resnet} we compare the overdamped variant o-CoLod (with $\tau = 0$) to its unconstrained counterpart. We observe that the use of an orthogonality constraint gives lower test loss throughout training.


\subsection{Circle constraints}\label{tableFashionMNIST}
For the results shown in Figure \ref{overviewfig_CNet}, Figure \ref{avg_overviewfig_CNet}, Figure \ref{gradients}, and Table \ref{curvature} the first class of the data set is generated using 
\begin{align}
 x& = \sqrt{t} \cos(4 \sqrt{t} \pi) +0.05\mathcal{N}(0,1), \nonumber \\
 y& = \sqrt{t} \sin(4 \sqrt{t} \pi) +0.05\mathcal{N}(0,1), \label{spiraleqn2turn}
\end{align}
where $t$ is repeatedly drawn from $\mathcal{U}(0,1)$. The other class is obtained by shifting the argument of the trigonometric functions by $\pi$. For our experiments we used 100 training data points, 2000 test data points and 2\% subsampling. We use a 500-node single hidden layer perceptron, with ReLU activation and BCE loss. We choose the optimal weight decay value for SGD through line search. The results in Fig. \ref{gradients} were obtained by computing the gradient of the predictions of a trained classifier (after 10,000 epochs) on a 1000x1000 grid using second order accurate central differences. 

For our Fashion-MNIST \cite{FashionMNIST} example we reduce the number of training data samples to 10,000 and we increase the number of test data samples to 60,000. We use a 1000-node SHLP with ReLU activation, cross entropy loss and batchsize 128. Our main result with our circle constrained approach is presented in Figure \ref{CNet_FashionMNIST}, the accompanying mean test accuracies with standard deviations are: 87.63$  \pm$0.04\% (c-CoLud), 87.39 $\pm$0.06\% (SGD), 87.47 $\pm$0.38\% (SGD with WD = 1e-4), 87.29 $\pm$0.58\% (SGD with WD = 5e-5), 87.45 $\pm$0.06\% (SGD with WD = 1e-5). Hyperparameters SGD: $h = 0.1, mom = 0.8$. Hyperparameters c-CoLud: $h = 0.3, \gamma = 1, r_0 = 0.05, r_1 = 0.1, \tau = 0$.

\begin{figure*}[!h]
\centering
\includegraphics[clip, trim=0cm 5cm 0cm 5cm,width=0.4\linewidth]{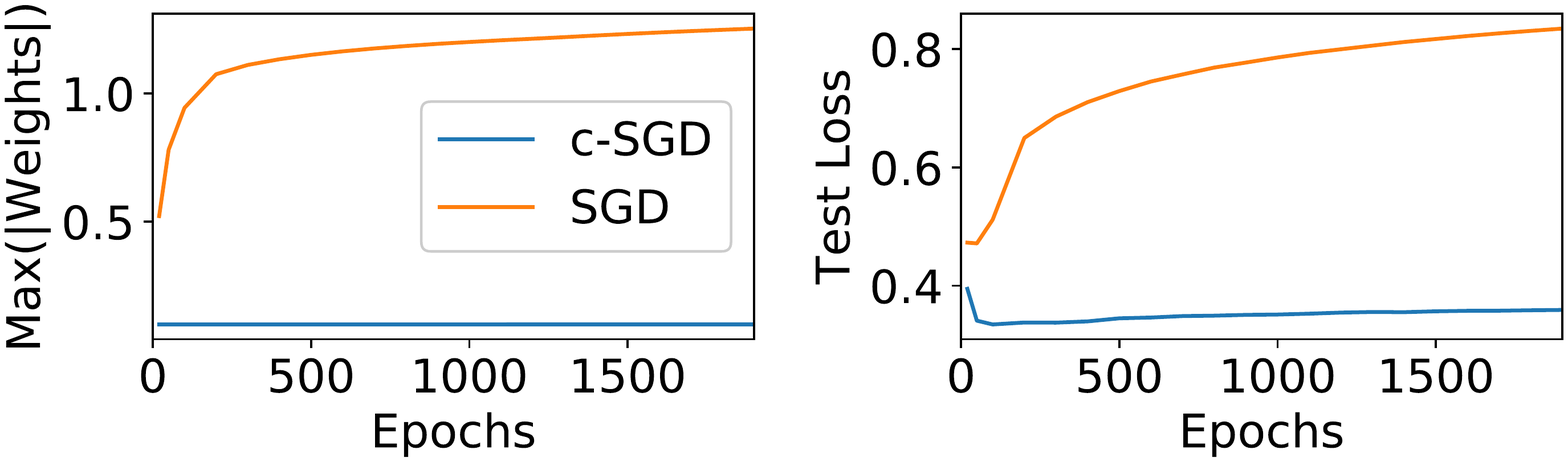}
\caption{Result is obtained for the Fashion-MNIST dataset with the same hyperparameter settings as in Figure \ref{CNet_FashionMNIST} of the main paper. We observe that the maximum absolute size of weights in the output layer of the network (left) and test loss (right) remain small and stable throughout training for the circle constrained method (c-SGD or c-CoLA-ud). In contrast SGD shows clear signs of overfitting.}\label{maxweightsCNet}
\end{figure*}

In Table D\ref{FashionMNIST_WD} we present extensive hyperparameter tests for the test accuracy and test loss obtained after 400 epochs (averaged over 5 runs) using SGD-m with and without weight decay (WD). In Figure D\ref{maxweightsCNet} we show that both the test loss and the maximum magnitude of the weights of the network remains small and stable throughout training for our circle constrained approach, while SGD shows signs of overfitting.

We also evaluate the performance of a small transformer model \cite{Transformer} on the Penn Treebank \cite{PennTreebank} and Wikitext-2 \cite{Wikitext2} data. The transformer has 2 encoder layers. Each encoder layer consists of self-attention with 2 heads and a feedforward network with 200 nodes followed by layer norms. We use batchsize 1024 for the Penn Treebank data and batchsize 128 for the Wikitext-2 dataset. We present the lowest validation loss obtained in 200 epochs by SGD-m and our circle constrained method c-CoLud in Table \ref{NLPexp} of the main paper. In Table D\ref{NLPexpwithWD} we provide a comparison with weight decay. 
 
 \makeatletter
\renewcommand{\fnum@table}{\tablename~D\thetable}
\makeatother
  \begin{table}[!hb]
 \vskip -0.1in
\caption{Minimum validation loss on Penn Treebank and Wikitext-2 using a transformer trained using SGD-m. We found weight decay set to WD = 1e-4 to give the best results for SGD-m. In comparison, the transformer trained using c-CoLA-ud obtains a minimum validation loss of 4.81 (Penn Treebank) and 5.09 (Wikitext-2). Using c-CoLA-ud therefore outperforms standard SGD-m in the case without WD, but does less well than SGD-m with weight decay, if the magnitude of the weight decay has been carefully tuned. In contrast, for Fashion-MNIST image data using a MLP we find that c-CoLA-ud outperforms both SGD-m with weight decay and SGD-m without weight decay (see Table D\ref{FashionMNIST_WD}).} \label{NLPexpwithWD}
\center
\begin{tabular}{c|c|c} 
      \textit{Optimizer} & Penn Treebank & Wikitext-2 
      \\ \hline 
     Without WD \ \ \ \ \ \ \ \ \ \ \  & & \\
  \ \ \ \ \ \ \ \ \ $mom$ = 0.7 & 4.87 & 5.13  \\
  \ \ \ \ \ \ \ \ \ $mom$ = 0.8 & 4.83 & 5.13  \\
  \ \ \ \ \ \ \ \ \ $mom$ = 0.9 & 4.84 & 5.13  \\ \hline
  With WD \ \ \ \ \ \ \ \ \ \ \  & & \\
  \ \ \ \ \ \ \ \ \ $mom$ = 0.7 & 4.84 & 5.01  \\
  \ \ \ \ \ \ \ \ \ $mom$ = 0.8 & 4.77 & 5.02  \\
  \ \ \ \ \ \ \ \ \ $mom$ = 0.9 & 4.77 & 5.02  \\ 
\end{tabular} 
\end{table}

\begin{table*}[!h]
\caption{These results are obtained for the Fashion-MNIST dataset using SGD with momentum to train a 1000-node SHLP. The results presented in the two right-hand columns are all obtained with weight decay set to 1e-4. We found this value to give the best results for SGD-m during a hyperparameter search.  
In comparison to the results for SGD-m shown in this table our circle constrained net reaches test accuracy \textbf{87.63\%}, with test loss \textbf{0.386} without using weight decay (see Figure \ref{CNet_FashionMNIST}). Hence it outperforms standard SGD with momentum both with and without weight decay. }\label{FashionMNIST_WD}
\center
\begin{tabular}{c|c|c|c|c|c}
  \multicolumn{2}{c|}{} & \multicolumn{2}{c|}{no WD} & \multicolumn{2}{c}{with WD} \\ 
    \multicolumn{2}{c|}{SGD with mom} & Test Acc. & Test Loss & Test acc. & Test Loss  \\ \hline
    h = 0.2 & mom = 0.8  & 87.18\% & 1.06 & 84.05\% & 0.696 \\
    & mom = 0.7  & 87.38\% & 0.890 & 87.0\% & 0.547 \\
     \hline
    h = 0.1  & mom = 0.9 & 86.97\% & 1.133 & 85.35\% & 0.634 \\ 
    & mom = 0.8 & 87.39\% & 0.824 & 87.47\% & 0.531 \\
             & mom = 0.7 & 87.39\% & 0.750 & 87.25\% & 0.517 \\ \hline
    h = 0.05 & mom = 0.95 & 86.67\% & 1.226 & 85.63\% & 0.623\\
    & mom = 0.9 & 87.33\% & 0.837 & 86.24\% & 0.569 \\
             & mom = 0.8 & 87.27\% & 0.719 & 87.33\% & 0.511\\
\end{tabular}
\end{table*}

\newpage
 \newpage \ \ \ \ 
  \begin{figure*}[ht]
     \centering
     \includegraphics[clip, trim=0cm 3.5cm 0cm 3.5cm,width=0.8\linewidth]{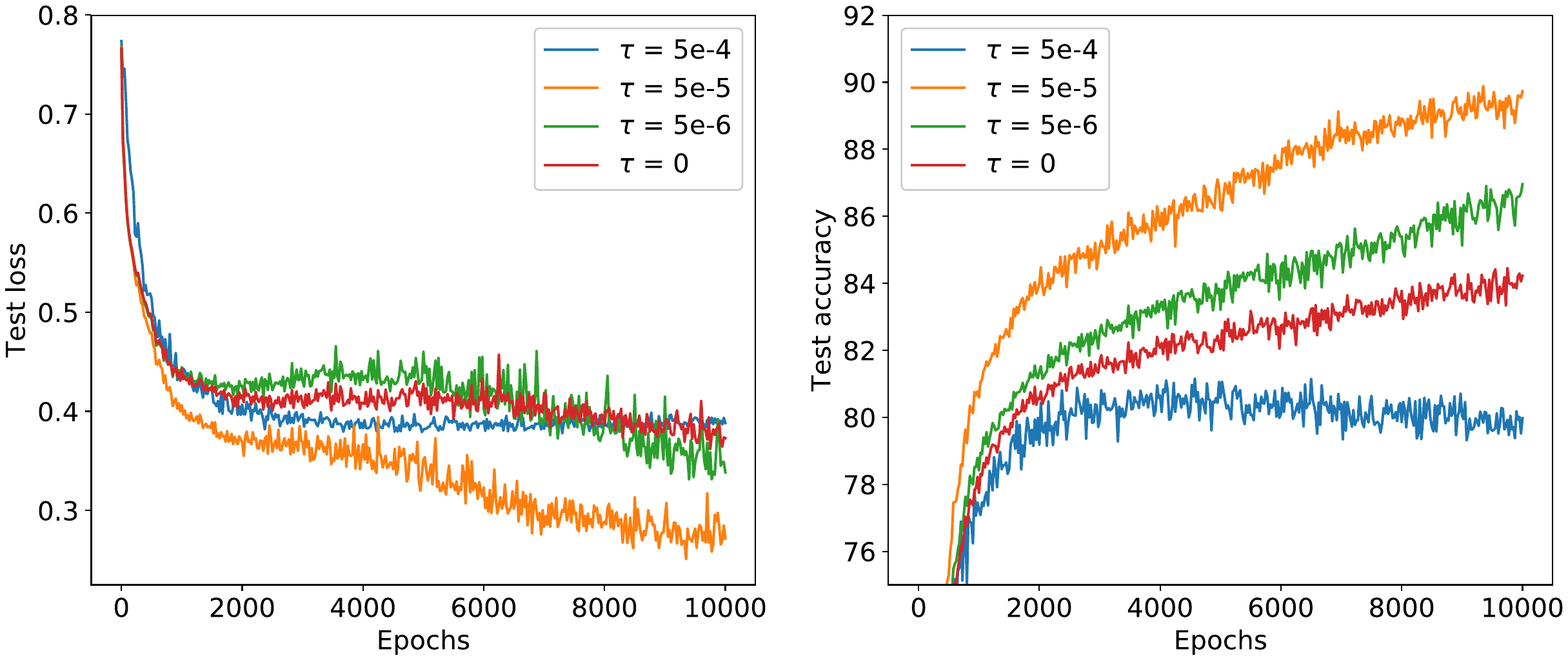}
     \caption{Test loss and test accuracy averaged over 100 runs for constrained approaches with varying levels of additive noise, i.e., with different values of the temperature hyperparameter $\tau$. The set-up is the same as for Fig. \ref{overviewfig_CNet} and Fig. \ref{avg_overviewfig_CNet} in the main paper, i.e., we train a 500-node single hidden layer perceptron for a spiral binary classification problem (Eq. \eqref{spiraleqn2turn}). Hyperpar. settings: $h = 0.05$, 2\% subsampling, $r_0 = 1, r_1 = 5$  
(see Eq. (\ref{eq:circleConstraint})). The best performance is obtained using temperature $\tau$ = 5e-5. This is also the temperature that results in the classifier with the lowest curvature estimate (see Table D\ref{curvature_TdependenceCSGLD}). }\label{TdependenceCSGLD}
 \end{figure*} 

\subsection{Curvature} 
It is difficult to establish a commonly agreed definition of curvature for a boundary that is potentially non-differentiable at a finite number of points. We computed our curvature estimates using the method described below which we suggest is indicative of the curvature of the locally smoothed classification boundary and allows us to compare the relative curvature estimates of classifiers trained using different optimizers. 
 
We evaluate the smoothness properties of our trained classifiers after a fixed number of 10,000 epochs. The curvature of a level curve $\phi(x,y)=0$ is defined as \cite{levelsetmethod}:
\begin{align*}
    \kappa = \nabla \cdot \frac{\nabla\phi}{|\nabla\phi|} = \frac{\phi_{xx}\phi^2_y-2\phi_y\phi_x\phi_{xy}+\phi_{yy}\phi^2_x}{(\phi_x^2+\phi_y^2)^{3/2}}
\end{align*}
 
However, since we do not have access to the exact form of $\phi$, we fit a contour to the model's predictions on a 1000x1000 grid using matplotlib.pyplot.contour,
which returns an array containing the coordinates of points along the contour. We view these as discrete samples from the parametric curve $(x(t),y(t))$. The gradients of these are computed using second order accurate central differences. This can then be used to compute the approximate curvature \cite{gray2006}:
\begin{align}
    \kappa = \frac{|x''y'-x'y''|}{(x'^2+y'^2)^{3/2}}
\end{align}
Although this results in a rough estimate, by averaging our results over 100 runs, we suggest this gives us some insight on relative curvature estimates of classifiers trained using different optimizers.

In Table D\ref{curvature_TdependenceCSGLD} and Figure D\ref{TdependenceCSGLD} we study the effect of varying the temperature hyperparameter $\tau$, which controls the additive noise level (see Eq. (\ref{eq:constrainedOverdampedLangevin})), on the generalization performance and curvature of the resulting classifiers on the spiral dataset defined by Eq. \eqref{spiraleqn2turn}. We show that there appears to be an ideal choice of temperature (in this case $\tau$ = 5e-5), for which the best generalization performance is obtained using our circle constrained approach. We also show that the trained classifier which has the lowest curvature estimate also obtains the best generalization performance.

 \begin{table}[!ht]
 \makeatletter
\renewcommand{\fnum@table}{\tablename~D\thetable}
\makeatother
\caption{Same set-up as for Fig. \ref{overviewfig_CNet}, \ref{avg_overviewfig_CNet}, and Figure D\ref{TdependenceCSGLD}. We present the mean curvature, standard deviation (std), and maximum (max) curvature of classifier boundaries obtained using our constrained approach with different values of the temperature $\tau$. The lowest curvature is obtained using $\tau$ = 5e-5, which also corresponds to the classifier which obtains the best generalization performance (see Figure D\ref{TdependenceCSGLD}). These results are averaged over 100 runs.} \label{curvature_TdependenceCSGLD}
\center
\begin{tabular}{c|c|c|c} 
     &  \multicolumn{3}{c}{\textit{Curvature Approximation}} \\ 
      \textit{$\tau$} for C-SGLD & Mean & Std & Max
      \\ \hline 
    $\tau$ = 0 & 9.38 & 317 & 5.58 $\cdot 10^5$ \\ \hline
    $\tau$ = 1e-6 & 9.01 & 273 & 1.63 $\cdot 10^6$ \\ \hline
      $\tau$ = 5e-6 & 7.75 & 166 & 5.86 $\cdot 10^5$ \\ \hline
       $\tau$ = 1e-5 & 7.06 & 108 & \ 4.06 $\cdot 10^5$ \\ \hline
       $\tau$ = 5e-5  & 6.08 & 40.8 & \ 1.43 $\cdot 10^5$ \\ \hline
      $\tau$ = 1e-4 & 7.62 & 178 & \ 9.47 $\cdot 10^{5}$  \\ \hline     
      $\tau$ = 5e-4 & 15.9 & 850 & \ 5.07 $\cdot 10^{6}$ 
\end{tabular} 
\end{table}

\end{document}